\documentclass[journal ]{new-aiaa}
\usepackage[utf8]{inputenc}
\usepackage{textcomp}

\usepackage{graphicx}
\usepackage{amsmath}
\usepackage{physics} 
\usepackage[version=4]{mhchem}
\usepackage{siunitx}
\usepackage[table]{xcolor} 
\usepackage{longtable,tabularx}
\usepackage{textcomp} 
\usepackage{tikz} 
\usepackage{booktabs} 

\makeatletter
\def\@fnsymbol#1{\ensuremath{\ifcase#1\or 
*\or 
\dagger\or 
\ddagger\or   
\mathsection\or 
\mathparagraph\or 
\|\or 
**\or 
\dagger\dagger\or 
\ddagger\ddagger\or 
\mathsection\mathsection\or \mathparagraph\mathparagraph\or 
\|\|\or
***\or 
\dagger\dagger\dagger\or 
\ddagger\ddagger\ddagger\or \mathsection\mathsection\mathsection\or \mathparagraph\mathparagraph\mathparagraph\or 
\|\|\|\or
\else\@ctrerr\fi}}
\makeatother 

\makeatletter
\renewcommand\@makefntext[1]{\leftskip=2em\hskip-2em\@makefnmark#1}
\makeatother 

\definecolor{LightBlue}{rgb}{0.88,0.88,1}

\title{Flight Demonstration and Model Validation of a Prototype Variable-Altitude Venus Aerobot}

\author{
Jacob S. Izraelevitz\footnote{Flight Test Principal Investigator and Robotics Technologist, Extreme Environment Robotic Systems Group. AIAA Member. 

Corresponding author:  jacob.izraelevitz@jpl.nasa.gov}, 
Siddharth Krishnamoorthy\footnote{Research Technologist, Ionospheric and Atmospheric Remote Sensing Group.},
Ashish Goel\footnote{Robotics Technologist, Robotic Surface Mobility Group.},
Carolina Aiazzi\footnote{Robotics Technologist, Robotics Modeling and Simulation Group; currently Graphics Engineer at Surreal Things, Inc.},
Michael Pauken\footnote{Group Supervisor, Advanced Thermal Concepts and Analysis Group.},
Gerald Walsh\footnote{RF Microwave Engineer, Exciter and RF/Millimeter Instruments Group.},
Carl Leake\footnote{Robotics Technologist, Robotics Perception Systems Group; currently Senior Technical Staff at Karana Dynamics.},
Christopher Lim\footnote{Robotics Software Engineer, Robotic Software Systems Group (retired).},
Abhinandan Jain\footnote{Senior Research Scientist, Robotics Section Staff; currently Chief Technologist, Karana Dynamics. },
Leonard Dorsky\footnote{Optical Engineer, Optics Section.},
Kevin H. Baines\footnote{Scientist, Planetary and Exoplanetary Atmospheres.},
James A. Cutts\footnote{Program Area Manager, Technology Program Office, Planetary Science Directorate (retired).},
Jeffery L. Hall\footnote{Deputy Manager, Technology Program Office, Planetary Science Directorate. AIAA Associate Fellow.}
}
\affil{Jet Propulsion Laboratory, California Institute of Technology, Pasadena, CA., 91109, USA.}

\author{
Caleb Turner\footnote{Senior Aerospace Engineer, Tillamook Site.},
Carlos Quintana\footnote{Director of Engineering, Tillamook Site; currently System Engineering Consultant, Blue Crane Engineering},
Kevin Carlson\footnote{Production Manager, Tillamook Site.},
Tim Lachenmeier\footnote{Lead Aerospace Engineer and Flight Range Safety Officer, Tillamook Site.}}
\affil{Aerostar LLC, Tillamook, OR, 97141, USA.}

\author{
Paul K. Byrne\footnote{Associate Professor of Earth, Environmental, and Planetary Sciences.}
}
\affil{Washington University in St. Louis, MO, 63130, USA.}

\begin{document}
\maketitle

\begin{tikzpicture}[overlay, remember picture]
\node[anchor=south west, xshift=1in, 
yshift=0.85in] at (current page.south west) {\copyright 2025 California Institute of Technology. Accepted document version.
};

\node[anchor=south west, xshift=1in, 
yshift=0.7in] at (current page.south west) {Published in AIAA Journal of Aircraft (JA) 2025. DOI: https://doi.org/10.2514/1.C038314
};
\end{tikzpicture}

\begin{abstract}
This paper details a significant milestone towards maturing a buoyant aerial robotic platform, or aerobot, for flight in the Venus clouds. We describe two flights of our subscale altitude-controlled aerobot, fabricated from the materials necessary to survive Venus conditions. During these flights over the Nevada Black Rock desert, the prototype flew at the identical atmospheric densities as 54 to 55 km cloud layer altitudes on Venus. We further describe a first-principle aerobot dynamics model which we validate against the Nevada flight data and subsequently employ to predict the performance of future aerobots on Venus. The aerobot discussed in this paper is under JPL and Aerostar development for an in-situ mission flying multiple circumnavigations of Venus, sampling the chemical and physical properties of the planet's atmosphere and also remotely sensing surface properties.


\medskip 
\end{abstract}

\pagebreak
\section*{Nomenclature}
{\renewcommand\arraystretch{0.6}
\noindent\begin{longtable*}{@{}l @{\quad=\quad} l@{}}
$A$ & area, m\textsuperscript{2} \\
$b$ & slope of pressure difference over ZP envelope, Pa/m \\
$C_d$ & vent discharge coefficient \\
$C_D$ & drag coefficient \\
$C_m$ & virtual mass coefficient \\
$c_v$ & constant volume heat capacity of helium gas, J/K \\
$c_p$ & constant pressure heat capacity of helium gas, J/K \\
$c$ & heat capacity of balloon material at node, J/K \\
$d$ & vent orifice diameter, m \\
$E$ & irradiance, W/m\textsuperscript{2} \\
$F$ & force, N \\
$f$ & view factor \\
$g$ & gravitational acceleration, m/s\textsuperscript{2}\\
$h$ & enthalpy of helium gas entering control volume, J \\
$m$ & mass, kg \\
Nu & Nusselt number of thermal node, either forced or natural  \\
$P$ & pressure, Pa \\
$Q$ & heat, J \\
$q$ & normalized meridional stress, N\textsuperscript{-1} \\
$q_i$ & net heat flux into thermal node $i$, J \\
$r$ & radius of ZP balloon shape from central axis at $s$, m\\
$R$ & radius of SP balloon from its center, m \\
$\overline{R}$ & specific gas constant, J/kg-K \\
Re & Reynolds number of thermal node and surrounding gas \\
Ra & Rayleigh number of thermal node and surrounding gas \\
$s$ & arclength of ZP balloon shape measured from bottom, m \\
$T$ & temperature, K \\
$t$ & time, s \\
$V$ & volume of helium gas, m\textsuperscript{3} \\
$v$ & velocity of aerobot, m/s \\
$W$ & thermodynamic work, J \\
$w$ & areal weight of balloon material, N/m\textsuperscript{2} \\
$z$ & height from bottom of aerobot, m \\

$\alpha$ & solar absorptivity of balloon material \\
$\beta$ & angle from vertical where the ZP material separates from SP, radians \\
$\rho$ & gas density, kg/m\textsuperscript{3} \\
$\theta$ & angle from vertical, radians \\
$\sigma_m$ & meridional stress on ZP balloon film at $s$, N/m \\
$\epsilon$ & infrared emissivity of balloon material \\

\multicolumn{2}{@{}l}{}\\ 
\multicolumn{2}{@{}l}{\textbf{Subscripts}}\\
0 & initial value at start of balloon shape domain\\
atm & external atmosphere \\
buoy & buoyancy \\
diff & difference between ZP and external atmosphere \\
gas & lifting helium gas \\
infl & fully inflated geometry \\
l & length of balloon shape domain \\
payload & payload below the aerobot \\
pump & pump from ZP to SP balloon\\
p0 & location where pressure is exactly equal between ZP and atmosphere \\
ref	& reference, as in area \\
side & side projected area of the computed aerobot shape \\
SP & super pressure balloon\\
sun & location of sun, as in angle to zenith \\
solar & shortwave flux from sun and reflected ground albedo \\
infrared & longwave flux from sky, atmosphere, envelope material, and ground \\
tension & meridional tension in balloon material\\
total & total of both balloons, their helium, and the payload below\\
top & top projected area of the computed aerobot shape \\
vent & vent from SP to ZP balloon\\
virtual & virtual mass of surrounding air\\
ZP & zero-pressure balloon\\

\end{longtable*}}    
\section{Introduction} \label{Sec.Introduction}

Venus, our enigmatic neighboring planet of similar size and solar distance, has evolved into a world with very different conditions as Earth with a dense carbon dioxide atmosphere, a cloud layer composed of sulfuric acid aerosols, and surface temperatures above 460$^\circ$C. The processes that governed Venus's evolution from a rocky planet's primordial state, and how and when Venus and Earth diverged, remain poorly understood despite a history of Venus exploration over the last fifty years. While large landed missions, probes, and orbiters have improved (and are planned to dramatically improve) our understanding of the planet, many science questions remain that can only be addressed with a well-instrumented long duration vehicle flying in the clouds.
\begin{figure} [b!]
    \centering
    \includegraphics[width=0.95\textwidth]{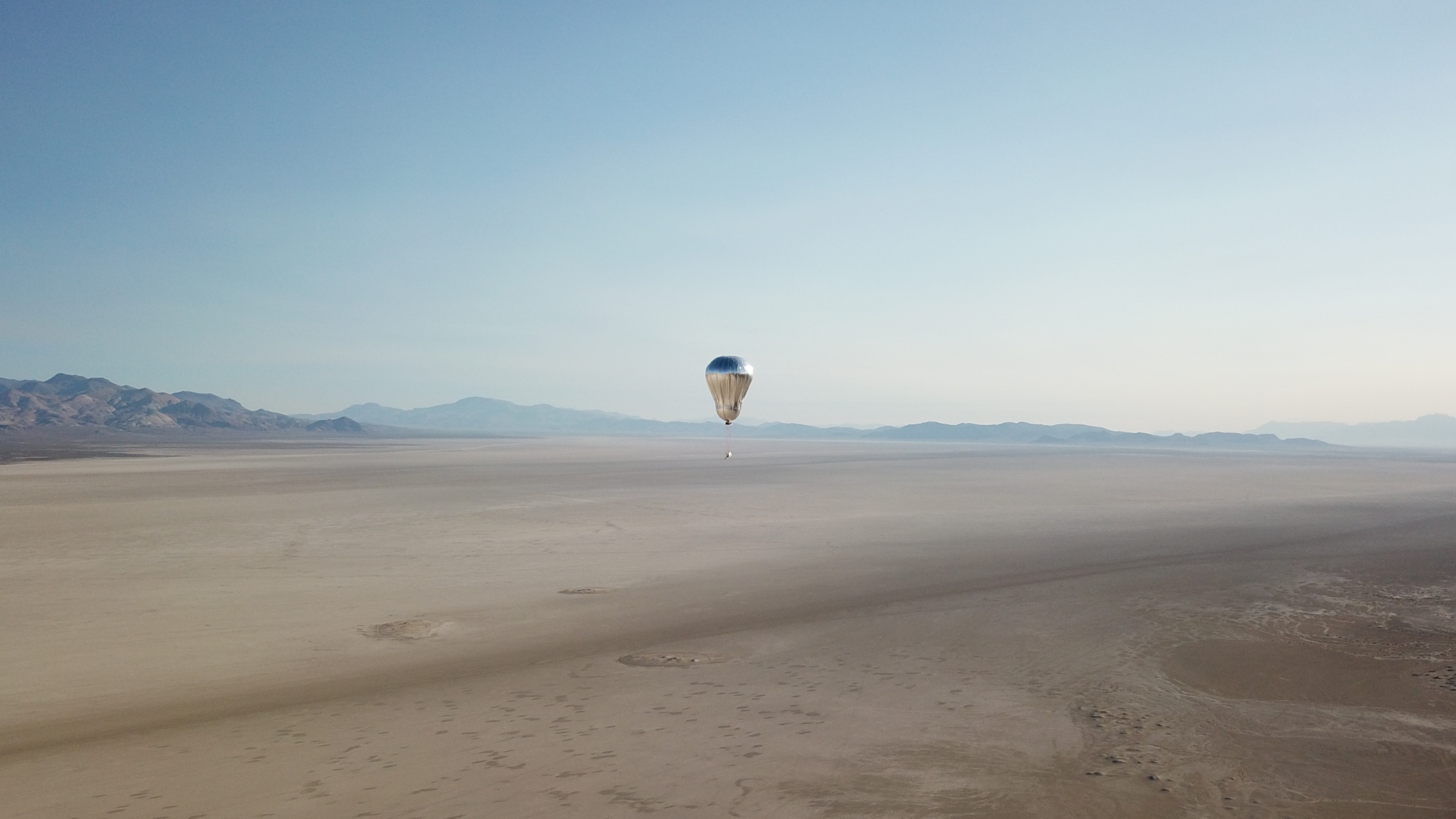}
    \caption{Subscale prototype of JPL's variable-altitude controlled aerobot, made from Venus-compatible materials, in flight over the Black Rock desert in Nevada, USA. Photograph taken July 2022, courtesy Aerostar. \copyright 2022 California Institute of Technology.}
    \label{fig:coverphoto}
\end{figure}
\subsection{Venus Aerobot Background}

Balloon experiments have operated on Venus before but none of the proposed capability. The Soviet VeGa lander missions of 1985 included balloon probes with two small 7~kg gondolas that flew for approximately 46~hours in the cloud layer at a nominally constant altitude \cite{marov2011soviet}. The challenge today is in expanding this early demonstrator to a larger and fundamentally more capable buoyant aerial robotic platform (aerobot), thereby enabling the greater science return promised by a dedicated mission. NASA's Jet Propulsion Lab (JPL) has been developing large aerobots for specifically this purpose for over twenty years, including multiple Discovery proposals \cite{balint2008nuclear}, carrying a 40-100~kg gondola below the aerobot with science instrumentation consistent with a much more complete characterization of the Venus cloud layer. These aerobots, like the VeGa balloons before, were designed to maintain altitude through the use of "superpressure" - i.e. a high strength envelope inflated to pressure slightly above the ambient atmospheric pressure.  Given its constant volume, a superpressure balloon loses buoyancy when displaced upwards and gains buoyancy when displaced downwards, making it altitude stable. 

More recently, a NASA study described in \citet{cutts2018venus} refocused JPL efforts on an aerobot that can instead actively control its altitude over a variable range (Figure \ref{fig:coverphoto}) instead of staying at a constant altitude. A variable-altitude approach allows the reuse of a single payload over a wide range of conditions in the clouds, thereby more fully characterizing the cloud layer (and providing an even greater science return) for a modest complexity cost. The success of the Google Loon program \cite{loon2021} provided credence to the rapid development of variable altitude aerobots for Earth, and the \citet{cutts2018venus} NASA study prioritized the adaption of this technology for Venus.  A number of variable-altitude Venus design points and their development are described in \cite{hall2021prototype}, with targeted altitude range on Venus from 52 to 62~km and a flight duration goal of 117 Earth-days (one Venus day). Such platforms are highly scaleable, with designs able to carry from 15~kg to 230~kg depending on mission ambition.

This paper describes a \textbf{significant milestone in the technology advancement of this Venus aerial plaform}: JPL's first outdoor flights of a prototype aerobot, made from Venus-compatible materials, at atmospheric densities within the proposed flight envelope for Venus. Table \ref{tab:atmos} describes the US Standard Atmosphere \cite{national1976us} that approximates the test flight conditions in comparison to the Venus atmosphere \cite{seiff1985models}. Balloon flight dynamics are principally driven by atmospheric density, meaning that this flight recreates a relevant environment for flight dynamics in Venus's cloud layer from 54-55~km above the surface. The targeted Venus temperatures are roughly $15 ^\circ$C greater than those in Nevada however, so models must be employed to take this and other environmental differences into account.  Accordingly, a key objective for this flight was to provide validation data for our flight dynamics model FLOATS (FLight Operations and Aerobot Trajectory Simulator), which we subsequently use to predict mission trajectories and operations on Venus. While flights at higher altitudes (up to 62 km equivalent density) are possible with our platform, balloons must trade altitude for payload mass for a given envelope volume, and a complete characterization of the balloon dynamics for FLOATS required more payload mass than the full altitude range could support on a prototype of this scale.

\begin{table}[t!]
    \centering
    \caption{Atmospheric equivalencies between Venus balloon altitudes and test flight altitudes (bolded rows) in Nevada, USA. Note exact density match, although Venus is slightly warmer.\label{tab:atmos}}
    \begin{tabular}{c c c c c c} \hline\hline
       \multicolumn{3}{c}{ \begin{tabular}{@{}c@{}} \textbf{Venus:} VIRA Reference \\ Atmosphere \cite{seiff1985models}\end{tabular}} & \multicolumn{3}{c}{ \begin{tabular}{@{}c@{}}\textbf{Earth:} US Standard Atmosphere\\ (incl. $+20^\circ$C local offset condition) \cite{national1976us} \end{tabular}} \\ \cmidrule(lr){1-3} \cmidrule(lr){4-6}
       Altitude & Density & Temperature & Altitude & Density & Temperature \\ 
       (km) & (kg/m\textsuperscript{3})& ($^\circ$C) & (km ASL) & (kg/m\textsuperscript{3}) & ($^\circ$C) \\ \hline 
       52 & 1.28 & 60.2 & -- & -- & -- \\ 
       53 & 1.15 & 49.9 & 0.0 & 1.15 & 35\\
       \textbf{54} & \textbf{1.03} & \textbf{39.7} & \textbf{1.1} & \textbf{1.03} & \textbf{27.9} \\
       \textbf{55} & \textbf{0.92} & \textbf{29.2} & \textbf{2.2} & \textbf{0.92} & \textbf{20.7} \\
       56 & 0.82 & 18.7 & 3.3 & 0.82 & 13.6\\
       57 & 0.72 & 9.4 & 4.5 & 0.72 & 5.8\\
       58 & 0.63 & 2.1 & 5.7 & 0.63 & -2.1\\
       59 & 0.54 & -4.4 & 7.0 & 0.54 & -10.5\\
       60 & 0.47 & -10.4 & 8.2 & 0.47 & -18.3\\
       61 & 0.41 & -14.5 & 9.4 & 0.41 & -26.1\\
       62 & 0.34 & -18.7 & 10.8 & 0.34 & -35.2\\ \hline\hline
    \end{tabular}
    \medskip
    
\end{table}

\subsection{Programmatic Context}

The development of this platform by JPL reflects preparation for a Venus mission concept that addresses the documented needs of the Venus science community. After assessing the state of fundamental knowledge on our sister planet, NASA's Venus Exploration Assessment Group (VEXAG) defined six overarching Venus science objectives that can be addressed through near-term Venus mission opportunities, such as "Did Venus have temperate surface conditions and liquid water at early times?" and "What processes drive the global atmospheric dynamics of Venus?" \cite{vexag2019}. While the recent selection of two Venus orbiters (VERITAS \& Envision) and the DAVINCI entry probe will go a long way towards fulfilling many of the objectives sought by the NASA Venus science community, the subsequent 2023-2032 Planetary Science and Astrobiology Decadal Survey Origins, Worlds, \& Life (OWL) notes that other Venus science objectives remain largely unaddressed by these missions, fundamentally because they require long duration in-situ measurements from within the atmosphere \cite{national2022origins}. OWL is the overarching guiding document for the scientific exploration of the planets for the next decade, so architectures that enable these long duration in-situ measurements will naturally be the focus of the next generation of Venus exploration.

Specifically, OWL has focused the New Frontiers 6 mission opportunity for Venus (called VISE, the Venus In-situ Explorer) on “the processes and properties of Venus that cannot be characterized from orbit or from a single descent profile" \cite{national2022origins}. These processes and properties include: "(1) Complex atmospheric cycles (e.g., radiative balance; chemical cycles, atmospheric dynamics, variations of trace gases, light stable isotopes, and noble gas isotopes, and the couplings between these processes); (2) Surface-atmosphere interactions (e.g., physical and chemical weathering at the surface, near-surface atmospheric dynamics, and effects upon the atmosphere by any ongoing geological activity); and (3) Surface properties (e.g., elemental and mineralogical composition of surface materials, heat flow, seismic activity, and any magnetization)” \cite{national2022origins}. 

Future missions to Venus can address VISE through several different architectures, including landers and aerial platforms, depending on the their focused science objectives.  While landed surface missions (at 460$^\circ$C) have their own distinct advantages and technology challenges, aerial platforms that operate in the cloud layer (60$^\circ$C to -20$^\circ$C) can also offer in-situ access within the atmosphere at much more clement thermal conditions - assuming the sulfuric acid aerosols can be held at bay. Specifically, \emph{buoyant} aerial platforms can avoid the hot surface conditions for the entire mission duration, thereby allowing the use of high-heritage instruments, avionics, and power systems, and creating a faster mission infusion opportunity with few thermal challenges \cite{cutts2018venus}.  Additionally, OWL has continued endorsing the Discovery program for concepts outside of VISE constraints, and Venus's nearby location in the solar system could make an ambitious aerobot mission as cost-effective as a less ambitious concept to farther planetary targets. 

The variable-altitude aerobot was accordingly adopted into two OWL-support mission concept studies for Venus \cite{gilmore2020venus,orourke2021advents}, and \citet{cutts2022explore} describe a variety of smaller missions that could be performed with a variable-altitude aerobot, including concepts for New Frontiers and Discovery proposals. The prototype flown in the flight tests described in this paper is approximately one-third diameter scale of the favored design point of these previous mission studies.




\subsection{Paper Outline}

Section \ref{sec:exp} describes the flight experiment parameters, including the subscale variable-altitude aerobot design, the approximately 30kg of instrumentation carried, and an overview of the flight operations and performance. Section \ref{sec:floats} describes the FLOATS simulation equations, an update since our prior description in \cite{hall2021prototype}, covering the balloon altitude dynamics, shape model, aerodynamics, thermodynamics, heat transfer, and atmospheric models. Section \ref{sec:results} compares the aerobot flight to the FLOATS reconstruction. Section \ref{sec:discussion} leverages FLOATS to predict flights on Venus, specifically a three-circumnavigation mission of a larger variable-altitude aerobot carrying a gondola of 100kg. Section \ref{sec:conclusion} concludes the paper with lessons-learned, next steps, and framing of the results within the larger mission context of aerial platforms for Venus.

A video of the test flights in this paper can be found at the media release accompanying the experiment at \href{https://www.jpl.nasa.gov/news/jpls-venus-aerial-robotic-balloon-prototype-aces-test-flights}{https://www.jpl.nasa.gov/news/jpls-venus-aerial-robotic-balloon-prototype-aces-test-flights}.

\section{Flight Experiment Design \label{sec:exp}}
\subsection{Aerobot Prototype Specifications}
The combined team at JPL and Aerostar have developed three subscale prototype variable-altitude Venus aerobots to-date, in increasing fidelity to the planned mission article, all fabricated in the Aerostar facility in Tillamook, Oregon. The prototype used in the flight tests in this manuscript is the second of the three and is the same aerobot as presented by \citet{izraelevitz2022subscale}. We restate some description of this aerobot here for context.

\begin{figure} [t!]
    \centering
    \includegraphics[width=\textwidth]{figures/launch.pdf}
    \caption{Photo of aerobot at takeoff, with the two balloon chambers (SP and ZP) labeled. Both balloons are made from the materials needed for a Venus mission, and helium is pumped/vented between balloons for altitude control. Photo courtesy NASA/JPL-Caltech, \copyright 2022 California Institute of Technology.}
    \label{fig:spzp}
\end{figure}

The aerobot is a balloon-in-balloon configuration (Figure \ref{fig:spzp}), with the ability to change buoyancy (and subsequently the altitude) by exchanging helium gas between the two balloon chambers carrying different gas densities \cite{voss2009advances}. The outer zero-pressure (ZP) balloon provides most of the buoyancy, while the inner superpressure (SP) balloon acts as a pressurized reservoir for helium gas and provides the remaining buoyancy. The SP balloon further maintains a stabilizing influence on the altitude dynamics of the balloon. This architecture, originally flown on Earth-specific climatological balloons by \citet{voss2009advances}, was found by \citet{hall2019altitude} to be the most mass-efficient and energy-efficient approach for variable altitude control when adapted to Venus. However, alternative variable altitude methods can be found in the literature; mechanically-compressed balloons can be more altitude stable on Venus \cite{de2015venus}, and the pumped-air architecture flown by Google Loon \cite{loon2021} are highly stable and low mass but is less suitable for Venus as sulfuric acid aerosols would be pumped into the balloon, necessitating internal protection measures. Passive cycling balloons have also been designed for Venus in the past \cite{nock1995balloon}, but this new generation of powered altitude control methods allows science teams to direct the aerobot to desired altitudes at specific longitudes or times of day for investigations.

A primary advantage of our chosen balloon-in-balloon architecture for Venus is the separation of functions for the two balloon envelopes: the outer ZP balloon envelope must survive the corrosive Venus atmospheric aerosols and intense sunlight but not high pressurization loads, while the inner SP balloon is designed for high pressurization loads while protected from environmental conditions by the ZP balloon. The separation of function allows for more mass-optimal materials for each envelope. Furthermore, the lack of significant pressure across the ZP envelope improves mission lifetime by not driving out helium gas through envelope defects.

Figure \ref{fig:laminate} illustrates the design of the two balloon envelopes. The outer ZP is a bilaminate of Teflon\textsuperscript{\textregistered} fluorinated ethylene propylene (FEP) and Kapton\textsuperscript{\textregistered} polyimide, with two metalization layers and a bonding adhesive. The Teflon\textsuperscript{\textregistered} offers excellent resistance to the clouds of sulfuric acid aerosols during Venus environment evaluations \cite{hall2008prototype}, while the Kapton\textsuperscript{\textregistered} provides a helium gas barrier and the tensile strength for carrying the payload. The silver metalization acts as a second surface mirror  (with a measured solar absorptivity $\alpha=0.08$ and infrared emissivity $\epsilon = 0.52$) to reduce solar heating of the helium, and second metalization layer adds extra protection to reduce the effect of helium diffusion through the material. While this specific material has not gone to permeability testing, a prior material of similar metalization construction showed < 5 cm\textsuperscript{3}/m\textsuperscript{2}-day as measured at room temperature \cite{hall2011technology} consistent with the desired lifetime on Venus.  The bilaminate design further adds resistance to pinholing from manufacturing defects, as the chances of two defects aligning between the layers is extremely small. The ZP balloon is constructed as a 60-degree sphere-cone-sphere with upper diameter of 5 meters and lower diameter of 2.5 meters.

\begin{figure}[t!]
\centering
\includegraphics[width=\textwidth]{figures/envelopes.pdf}
\caption{A: Material stackup of the zero-pressure (ZP) balloon bilaminate. B: Two layers of the super-pressure (SP) balloon. C: Photograph of SP balloon. Modified with permission from \citet{izraelevitz2022subscale}, \copyright2021 California Institute of Technology.}.
\label{fig:laminate}
\end{figure}

The inner SP balloon is fabricated as a 2.5~m diameter sphere, also two-layered, with a Vectran\textsuperscript{\textregistered} fabric  (200 denier 2x2 basket weave at 76 threads-per-inch) to provide pressurization strength and a heat-sealed thermoplastic urethane bladder for gas retention. The two layers are indexed together with attachment tabs to keep relative alignment. While the ZP material is metalized, the urethane need not be. Helium diffusion through the SP balloon is not lost to the atmosphere but goes into the ZP balloon and hence the only effect is to slightly increase the pumping load, which was found to be easily accommodated by the pumping rate during assembly-level testing \cite{hall2011technology}. The SP balloon was proof-tested to 14~kPa before the flight, and an identical SP balloon was taken to 61~kPa to provide confidence in the planned pressurization margin utilized in the flight tests. 

The SP balloon naturally sinks to the lower half of aerobot as the compressed SP helium is denser than the ZP helium. This balloon arrangement makes the aerobot peanut-shaped at low altitudes and sphere-cone-shaped at higher altitudes (see shape model in Section \ref{subsubsec:shape}). Generally for field deployments on Earth the aerobot is inflated from the ground, but an internal load line between the SP apex and ZP apex also allows for a vertically-oriented hanging inflation as desired for Venus \cite{gatto2024inflation}, where inflation during atmospheric descent is preferred over a launch from the hot surface.

\subsection{Instrumentation, Gas Transfer, and Communications Payload}

\begin{figure} [t!]
    \centering
    \includegraphics[width=0.65\textwidth]{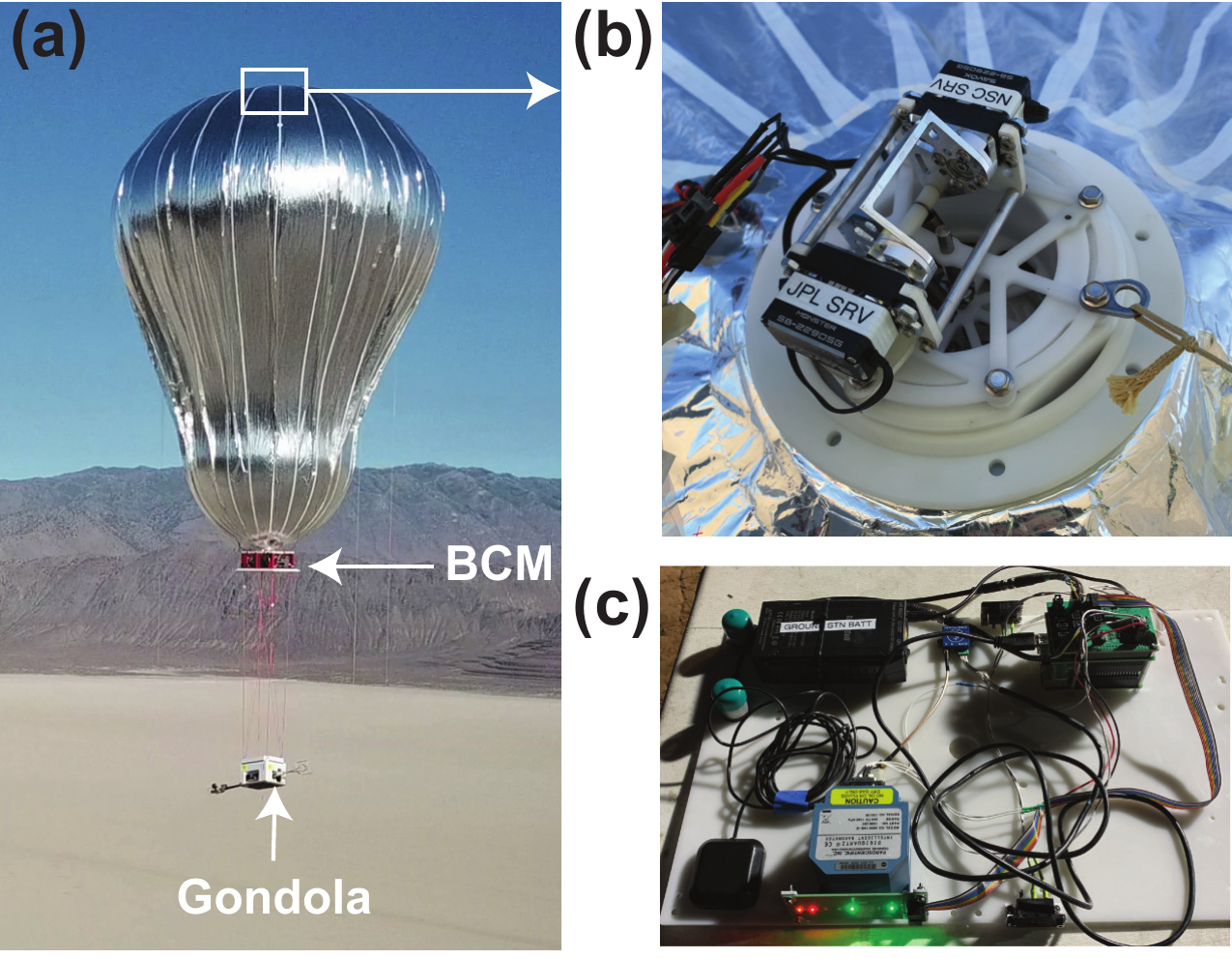}
    \caption{Instrumentation, gas transfer, and communication systems on the aerobot, divided into multiple platforms in different locations. (a) Buoyancy Control Module (BCM) and Gondola hanging below the aerobot. (b) Poppet for flight termination at the aerobot apex. (c) Ground station with stationary instrumentation.}
    \label{fig:bcmGondola}
\end{figure}

For accurate measurement of the flight dynamics of the aerobot, we designed and implemented an instrumentation system (Figure \ref{fig:bcmGondola}) that could not only gather information about the 6-DOF dynamics of the aerobot, but also on its thermal state and driving inputs from the environment such as atmospheric temperature, pressure, wind, and solar radiation. In addition, our instrumentation package coordinated the gas transfer between the two aerobot chambers, executed flight termination processes, and sent back real-time telemetry. This payload module was fabricated using hobby-grade avionics and commercial-of-the-shelf instrumentation for ease of development timeline, and would be replaced by spaceflight-grade systems for a Venus mission. This payload is a second iteration of what was flown in our previous indoor flights in Tillamook Airship Hangar \cite{izraelevitz2022subscale} with additions for wind measurements, solar forcing, long distance commanding, and Federal Aviation Administration (FAA) requirements. Components are summarized in Table \ref{tab:instrumentsTable}. Some sensors (such as the temperature sensors) are sampled much faster than the expected time scale of the dynamics of the measured quantities. This was done primarily for two reasons -- first, the presence of a large number of data points to average reduces the noise in the measurement, and second, a fast sample rate was well within system capabilities and allowed for an abundance of caution against any aliasing in a untested dynamical environment.

\begin{table}[t!]
\caption{\label{tab:instrumentsTable} Summary of Flight Test Payload}
\centering
\resizebox{\textwidth}{!}{
\begin{tabular}{lllc}
\hline \hline
Name & Type & Platform(s) & Sample Rate (sps) \\\hline
\underline{Flight Instrumentation} &  &  &  \\
Paroscientific 6000-15A-IS & Atmospheric absolute barometer  & BCM, Gondola, Ground & 50\\
InertialSense $\mu$INS & Inertial Navigation System (INS) with dual GPS receiver on 1m baseline & BCM, Gondola, Ground & 30\\
Texas Instruments LM35-CA/Z & Skin temperature sensor x4 and atmospheric temperature x1 & BCM & 20 \\
TE Connectivity 5525DSO-SB005GS & Differential barometer (vented gauge) & BCM & 10\\ 
Apogee Instruments SP510/610 & Pyranometers (solar radiation, up to 2 $\mu$m wavelength) x4 & Gondola, Ground & 30\\
Apogee Instruments SL510-SS/610-SS & Pyrgeometer (IR radiation 5--30 $\mu$m) x2 & Gondola, Ground & 30\\
Windmaster Pro & Ultrasonic anemometer for relative wind and atmos. temperature & Gondola & 20\\
Freshliance Temperature Logger & Internal gas temperature sensor & Poppet & 1/300\\
GoPro Hero 8 & Timelapse Cameras x2, sidelooking and uplooking & Gondola & 1\\
Inspeed Cup Anemometer & Groundspeed winds (display only, not logged) & Boom above transport trailer & 1 \\
\hline
\underline{Gas Transfer} &  &  &  \\
Boxer 3KQ 24V Diaphram & Fixed rate positive displacement helium pump x2  & BCM & --\\
PRM 1.4Cv 24V Solenoid  & Vent valve for SP-to-ZP helium flow, 7mm orifice & BCM & -- \\
Asco 4Cv 24V Solenoid  & Safety gate valve for pump flow, 14mm orifice & BCM & -- \\ 
Hitec HS-1005SGT  & Flight termination servos x2 & Poppet & -- \\\hline
\underline{Communication} &  &  &  \\
Uavionix PING-200SR 250W & ADSB transponder for air traffic control & Gondola & -- \\ 
NAL Research SAF5350-C & Iridium L-band antenna & Gondola & -- \\ 
Aerostar C3 Unit (Custom) & GNSS receiver, Iridium \& 902 MHz transceivers & BCM & -- \\
JPL Avionics Board  (Custom) & Raspberry Pi 4 datalogger, power distribution, PWM generator & BCM, Gondola, Ground  & -- \\
\hline \hline
\end{tabular}
}
\end{table}

\subsubsection{Payload Platforms}

Conflicting placement requirements imposed by the variety of functions performed by the payloads required components to be partitioned into four different platforms, described below:  
   
\begin{enumerate}
    \item \textbf{Poppet Valve}: The poppet valve was located at the apex of the ZP balloon, allowing for rapid helium evacuation to terminate the flight. Two independent servos on separate power systems could open the poppet for dual redundancy. The mechanical design consisted of a custom 95~mm (3.75~inch) diameter orifice, preloaded shut with two tension springs, with the servos overcentered to release the preload springs on command. A temperature sensor additionally hung from the poppet into the ZP gas. We note that a termination poppet is not strictly required for a nominal Venus flight, but it could be paired with ballast drops to add additional emergency margin against vertical winds if the control authority of the gas transfer system was unexpectedly insufficient. The total balloon system mass was 21.3~kg, which includes the two balloon envelopes, their apex fittings, and the poppet valve subsystem.
    
   \begin{figure} [t!]
    \centering
    \includegraphics[width=0.8\textwidth]{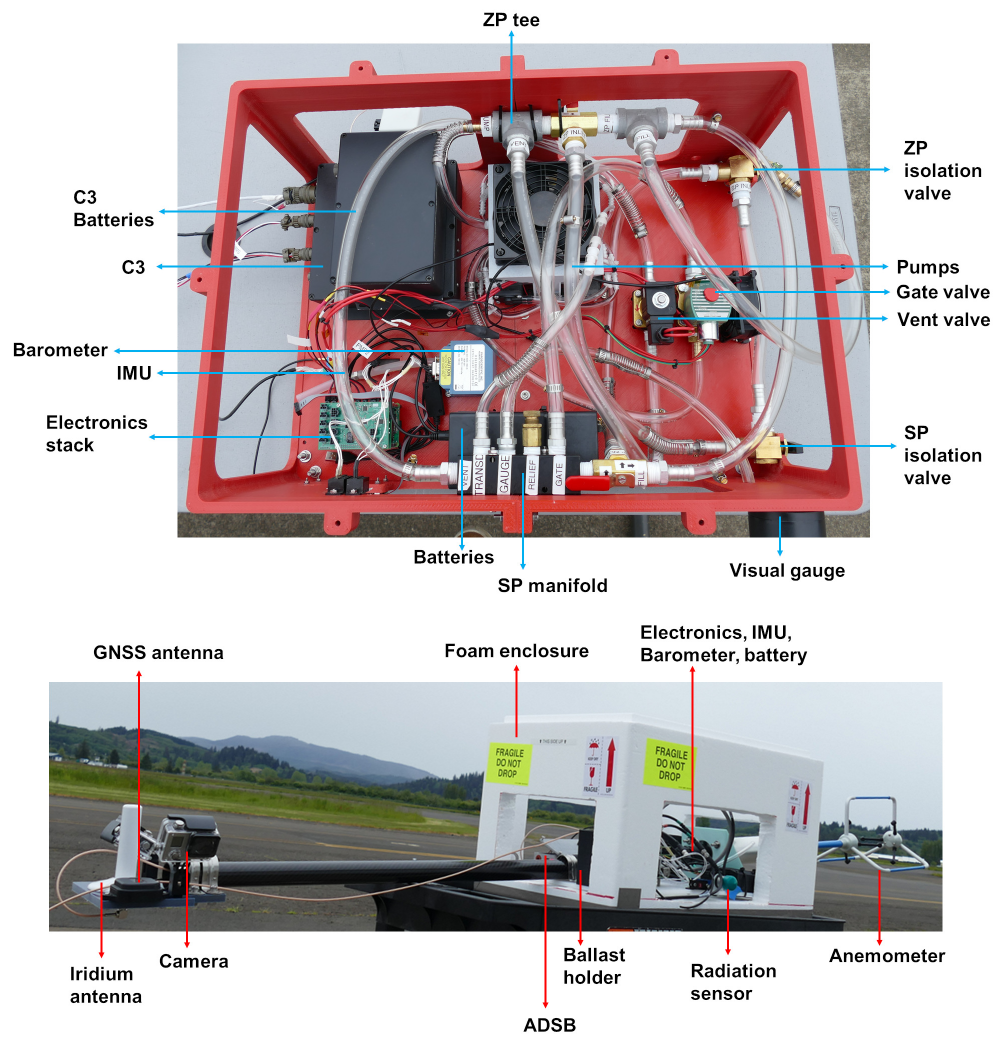}
    \caption{Annotated pictures of the buoyancy control module (top) and gondola (bottom), with various mounted instruments.}
    \label{fig:BCMGondolaDetailed}
    \end{figure}
    
    \item \textbf{Buoyancy Control Module (BCM)}: The BCM contained the entire gas transfer system, including a 12.7 mm (1/2 inch) inner diameter bypass tube through the entire vertical extent of the SP balloon to transfer gas into the ZP balloon. The BCM system needed to be as close to the balloon chambers as possible to reduce additional flow path length and associated loss of pressure; it was therefore fastened to the bottom end-fitting of the aerobot and would likely be further integrated into this end-fitting in a final mission configuration. The mechanical design centered around accommodating the volumes and masses of the C3 (Command, Control, and Communications) unit, pumps, valves, avionics, and batteries. A 3D-printed chassis was used to minimize structural mass and an aluminum honeycomb sandwich panel was used to interface with the bottom end-fitting of the balloons. Cables from the BCM to the flight termination poppet and skin temperature sensors ran along the exterior of the balloon for easy access, adhered to the envelope with FEP tape and covered with ZP material scraps to insulate against solar radiation. Figure \ref{fig:BCMGondolaDetailed}a shows annotated pictures of the BCM; the total mass of the BCM was 19.5~kg.
    
    \item  \textbf{Gondola}: The gondola contained instruments that needed to be separated from the aerobot and was suspended approximately 3 meters below the BCM. Incidentally, this type of sensor arrangement also replicates an expected partitioning of instruments and flight systems on the Venus aerobot, where certain components need to be either near the balloon for access or isolated from aerodynamic or electromagnetic interference from the balloon (the metalized ZP envelope has an especially strong effect on radio signals). The mechanical design of this gondola was dictated by this separation distance, the wind-sensor length, and a meter-long boom. The boom was necessary for three purposes: (1) to give the Iridium antenna a clearer view of the sky around the metalized ZP balloon; (2) to give the upward looking GoPro camera a clear view of the side of the ZP balloon envelope; and (3) to provide a long baseline for the differential GNSS antennas. The antenna for the Iridium transceiver was connected via a long coaxial cable to the transceiver of the C3 module inside the BCM. The ADS-B transponder was suspended below the gondola, with the base of the gondola covered in copper tape to shield the electronics from radio frequency interference during ADS-B transmission. Figure \ref{fig:BCMGondolaDetailed}b shows annotated pictures of the gondola, and the total mass of the gondola was 7.0~kg.
    \item \textbf{Ground Station}: The ground station was placed near the launch site to house instrumentation that need not be located on the aerobot. The mechanical design was a single Delrin\textsuperscript{\textregistered} plate, with all components fastened to it for easy transport. It was placed on an outdoor folding table with access to the solar incident environment for the duration of all flights.
\end{enumerate} 

\subsubsection{Assembly-level Payload Features}
Measurements made on the gondola and ground station were saved on-board and retrieved after the flight test. Data collected on-board the BCM were stored on board at the native sampling rate (variable, but $>10$ Hz for most instruments), and relayed back to the ground station through a radio link at 1 Hz for redundancy and flight monitoring. The measurements, commanding, and communication on the three platforms was performed by an identical flight computer stack. This ensured interchangeability and redundancy in the event of any failure. The flight computer stack and avionics diagram are shown in the supplementary material, along with post-flight calibration data of some of the instruments.

As the gondola would be the first module to impact the ground on landing, it was encased in a styrofoam enclosure, and the wind sensor was mounted on a $15^\circ$ wedge slightly upwards from horizontal. A foam plate was also added to the bottom of the BCM for impact mitigation. 

The tethered suspension between the gondola and BCM comprised of a single strand of paracord looped multiple times across attachment points at the four corners of the BCM and the gondola, with the goal of increasing the torsional stiffness of the suspension system. Hardly any relative angular motion was observed between the gondola and BCM during flight. The friction between the lines and their restraints did however make it harder for the gondola to level off during flight, which might benefit from some redesign for future efforts. 
    



\subsection{Flight Experiment Summary}
Two outdoor flights in Nevada were carried out in the summer of 2022 over a span of three days with the objective of demonstrating controlled flight of our subscale aerobot prototype in Venus-relevant atmospheric density conditions. The flights lasted more than an hour each and reached a maximum altitude above sea level (ASL) of 2.8~km and 2.4~km respectively, with the launch ground elevation being 1.3~km ASL. In this section we describe the preparation for and execution of the outdoor free flights of the aerobot. 

The field team included nine attendees: five JPL (Jacob Izraelevitz, Siddharth Krishnamoorthy, Ashish Goel, Gerald Walsh, Michael Pauken) and four Aerostar (Tim Lachenmeier, Caleb Turner, Allen Dial, Richard Bauer).

\subsubsection{Launch Site Selection}
As discussed in Section \ref{Sec.Introduction}, a low altitude flight in Earth atmosphere serves as a good analog for flights in Venus atmosphere in the target altitude range. This being the first outdoor flights of the prototype Venus aerobot, one of the main criteria in choosing a field site was the need for being able to terminate the flight at any point without risking either damage to the one-of-a-kind prototype by ground scrub brush or an unsafe overflight of populated areas. The Black Rock desert in Nevada was chosen as it offers many kilometers of unobstructed, flat lakebed for a chase vehicle to follow the balloon with a high likelihood of safe and speedy recovery upon landing on predictable smooth terrain. Bureau of Land Management permissions were obtained prior to flight for use of the lakebed, and flight vehicle parameters were filed in advance as a Certificate of Waiver or Authorization (COA) with the FAA. FAA was again notified by phone before and after the flights, and tracked live by the Oakland Air Route Traffic Control Center (ZOA). 

\subsubsection{Mission Planning \label{subsec:planning}}
The specific launch site on the lakebed, launch date, and launch time relied heavily on wind forecasts and anticipated balloon trajectories. NOAA GFS 0.25 degree wind forecast data \cite{gfs} was used to estimate wind as a function of altitude and to simulate the trajectory of the aerobot for a nominal FLOATS altitude profile (see Section \ref{sec:floats}). As the GFS atmosphere has resolution of only 0.25 degrees in latitude and longitude, predictions do not necessarily take into account local topographical variations. The trajectory predictions were accordingly only treated as rough indicators of the general direction and distance that the aerobot would cover during its flight. The flight relied on the poppet to terminate the flight early in case of the trajectory heading towards undesired territory. While the safe zone on the lakebed was initially defined using Google Earth, a further refinement was performed by a scouting team driving along the periphery of the lakebed on the day before first launch.

The other aspect of wind forecast relevant to mission planning is surface winds. Inflating, trimming the buoyancy of the balloon, and recovering the aerobot during high winds is challenging and potentially damaging for the balloon envelopes. On most days of the campaign, the winds were lowest around sunrise. Our goal was to launch with surface winds below 2 m/s and recover with surface winds below 3.5 m/s based on expertise at Aerostar. However, a morning launch also has significant undesirable thermal implications on the balloon - creating the largest helium and atmospheric temperature variations between launch and `cruise' phases of the flight under solar heating. This translates to a risk of flying to higher-than-desired altitudes or ground recontacts. Keeping these factors in mind, we selected a launch time close to sunrise for both the launches but kept the target free-lift (buoyancy above neutral) to under 500~g, as ground recontact could be mitigated by the recovery teams. 

Fourteen T-bottles of helium (8.5m\textsuperscript{3} per bottle) were brought to Nevada for the two flights, with the expectation of six bottles per flight (and one tank of margin) to meet target the desired lift condition. All equipment was brought to and from the lakebed each morning from accommodations in Gerlach in a array of sport utility vehicles, a towed groundstation trailer, and a recreation vehicle for crew breaks.

\subsubsection{Flight Tests}
\begin{figure}
    \centering
    \includegraphics[width=\textwidth]{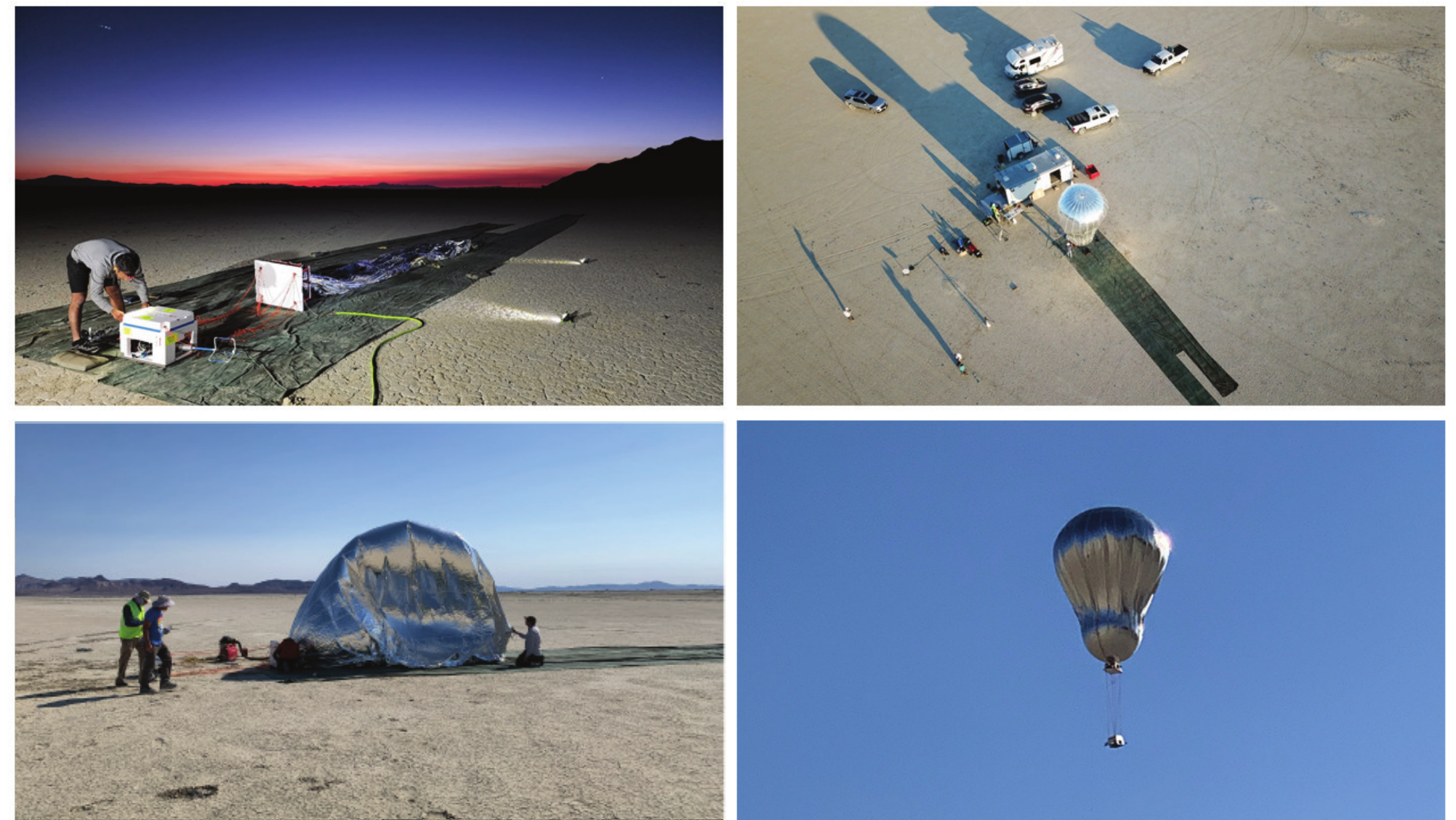}
    \caption{Clockwise from top left: equipment and aerobot being laid out on tarp prior to inflation (4:53AM); inflated aerobot prior to release (7:24AM); aerobot in free flight (7:33AM); aerobot being deflated after landing (9:28AM). Photographs from Flight 1, courtesy NASA/JPL-Caltech. \copyright 2022 California Institute of Technology.}
    \label{fig:FieldTestCollage}
    \end{figure}

Figure \ref {fig:FieldTestCollage} shows four images from the flight tests. Instrument checkouts were carried out a day before each launch. On the day of each launch, given the need to launch at sunrise, setup and operations began close to midnight. After a final integration and checkout of all the sub-systems, the SP balloon was inflated, followed by the ZP balloon, until the desired free-lift was obtained. This free-lift was measured and trimmed directly from the balloon buoyancy in the minutes before launch, as estimating indirectly from helium input (i.e. tank pressure or a mass flow meter) accumulates significant error with the large gas volume compared to the desired lift and is sensitive to the change in environmental temperatures over the hours between fill and launch.

The winds observed from the cup anemometer, especially during the first launch, were much higher than expected (in excess of 4 m/s) possibly due to topography near the lakebed boundaries. This made the launch operations more challenging and reduced our ability to accurately estimate the free-lift of the aerobot.  Consequently, in both the launches, the aerobot needed to be brought back to the ground, and team members in the chase vehicle had to re-ballast the payload gondola to achieve the desired free-lift condition. In the following sections of the paper, we primarily focus on the phases of the missions after the re-ballasting was done. 
    
During the flight, the pointable 902~MHz communication antenna instead of Iridium was primarily used to relay commands and telemetry as it was found to be more reliable, especially as the aerobot remained within line-of-sight. Once the aerobot reached the end of the safe operating zone after a few hours, the termination poppet was activated to release the helium from the ZP balloon and bring the aerobot to the ground. Simultaneously, the SP balloon was also vented to ensure that the SP balloon was not pressurized at the time of landing. There was sufficient helium left in the balloons to keep the system upright upon landing. The chase team was able to reach the aerobot within a minute of landing, and thereby prevent any damage to the aerobot envelope. The helium was then removed using both the termination poppet and a portable vacuum connected to the inlet ports to gently deflate the aerobot onto tarps set up by the chase team. 

\begin{figure}
    \centering
    \includegraphics[width=\textwidth]{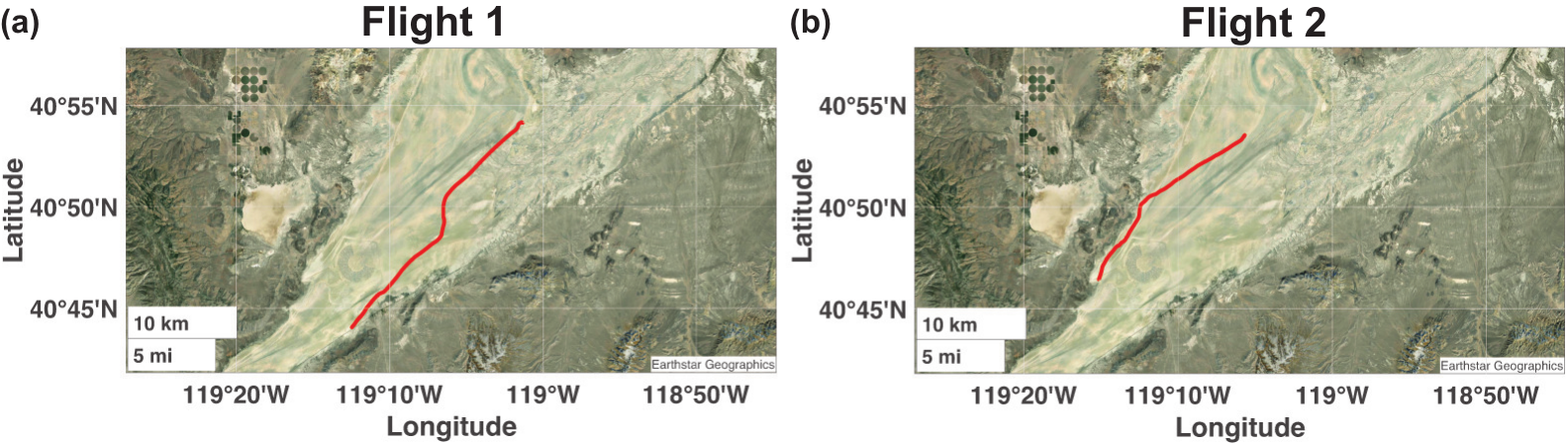}
    \caption{Ground track of (a) Flight 1 on July 14, 2022 and (b) Flight 2 on July 16, 2022, in the Black Rock desert. Flight launch locations were chosen based on wind forecasts and flights were terminated to ensure aerobot landing within the dry lakebed boundary.}
    \label{fig:GroundTracks}
    \end{figure}

Figure \ref{fig:GroundTracks} shows the ground tracks of the aerobot for the two flights. Note that in the second flight, the turn in the heading of the balloon was carried out through altitude adjustment by the ground operations team, to move the aerobot back towards the center of the lakebed based on predictions of how the wind varied with altitude. The wind predictions to inform this turn were obtained in real time by the pilot via GFS/ECMWF models from Windy.com \cite{windy2025}.

\section{FLOATS Simulation Model \label{sec:floats}}
The key objective of our two aerobot flights was to collect sufficient data for validation of our FLOATS simulations. This model bridges the gap between flights on Earth and flights under Venus conditions and is accordingly critical for the technology maturity of the aerobot. The following section describes the FLOATS model in detail, including the equations of motion of the dynamics, aerodynamics, balloon shape, thermodynamics, heat transfer, and external atmosphere.

\label{sec.FLOATS}
\subsection{FLOATS Ecosystem}
The FLOATS aerobot simulator is built on JPL's DARTS/Dshell toolkit \cite{darts}. The Dshell framework is designed to  develop complex  dynamics simulations that incorporate multibody dynamics, environmental models, and actuator/sensor device models for the closed-loop simulation of autonomous space vehicles, landers, and robotics applications. The reusable Dshell models are written in C++ for computational speed, and a Python interface is available for user interaction, configuration, and scripting support. FLOATS allows the selection of model configuration, environment, and design points from the command line as well as via configuration files, providing a convenient way to switch between Venus and Earth scenarios and different design points.



\subsection{Equations of Motion in FLOATS}
\subsubsection{Dynamics Model}
\begin{figure} [t!]
    \centering
    \includegraphics[width=0.8\textwidth]{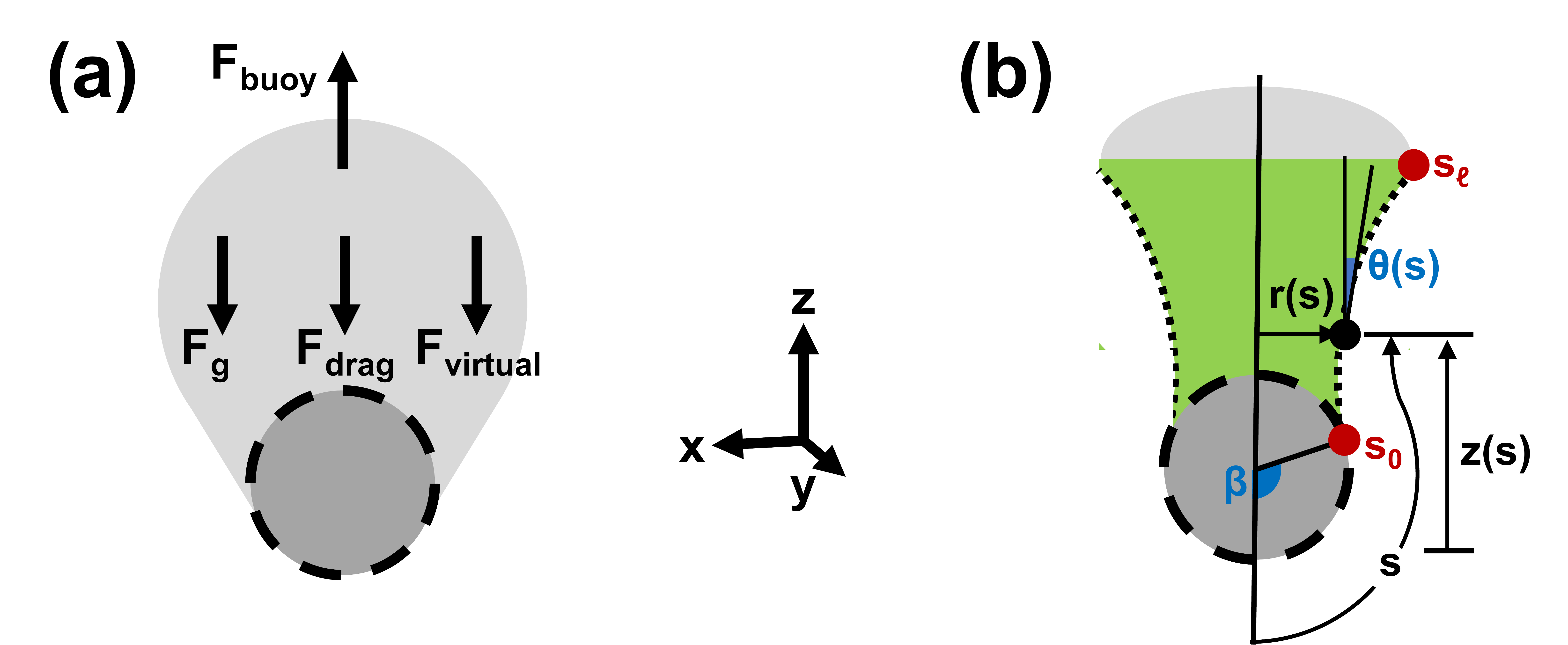}
    \caption{FLOATS variable definitions. (a) Vertical forces on aerobot assuming ascending velocity and acceleration (b) Shape model, where green mid section and its boundaries are solved by the computation.}
    \label{fig:force_cartoon}
\end{figure}

The principal dynamic equation governing the vertical motion of the aerobot system is presented in Eq. \ref{eq:eom}, which accounts for buoyancy, gravitational forces, drag, and virtual mass effects (Figure \ref{fig:force_cartoon}a).
\begin{align}
\label{eq:eom}
m_\text{total} \frac{d^2z}{dt^2} &= \sum F_z \nonumber \\ 
(m_\text{total} + m_\text{virtual}) \frac{d^2z}{dt^2} &= \rho_\text{atm} Vg - m_\text{total}g - F_\text{drag}
\end{align}
Where the displaced volume V is computed from the sum of the fixed SP balloon volume and the variable ZP balloon volume given by the ideal gas law. The horizontal motion is similarly governed (with buoyancy and gravity  neglected) and all masses lumped as points as the multibody swinging dynamics were not the focus of this experiment. Note that for this analysis, we itemize the total mass including helium gas as:
\begin{equation}
\label{eq:mass}
m_\text{total} = m_\text{SP,gas} + m_\text{ZP,gas} + m_\text{SP} + m_\text{ZP} + m_\text{payload}
\end{equation}
The FLOATS pump model computes the mass flow rate for the helium exchanged between the ZP and the SP balloons for adjusting the aerobot's buoyancy. The pump and the vent flow rates between the two balloons are computed with a constant-displacement pump relation and venting orifice relation of:
\begin{equation}
\label{eq:pump}
\dot{m}_\text{pump} = \rho_\text{ZP}  \dot{V}_\text{pump}
\end{equation}
\begin{equation}
\label{eq:vent}
\dot{m}_\text{vent} = C_d  \frac{\pi}{4}  d^2  \sqrt{2 \rho_\text{SP}  (P_\text{SP} - P_\text{ZP})  }
\end{equation}
The two net helium mass derivatives $ \dot{m}_\text{SP,gas}$ and $ \dot{m}_\text{ZP,gas}$ are then given as:
\begin{align}
\label{eq:dmsp}
\dot{m}_\text{SP,gas} = \dot{m}_\text{pump} - \dot{m}_\text{vent} \nonumber \\
\dot{m}_\text{ZP,gas} = \dot{m}_\text{vent} - \dot{m}_\text{pump}
\end{align}

\subsubsection{Shape Model \label{subsubsec:shape}} 
The balloon shape varies with time, and accordingly must be computed each timestep to determine the vertical drag area. The FLOATS shape computation assumes that the balloon shape is axisymmetric, is in a shape equilibrium, and is parameterized in terms of arc length. The shape takes a two-lobed path from the bottom apex to top apex of the ZP balloon (Fig. \ref{fig:force_cartoon}b), pushed towards and away from the central balloon axis as dictated by the circumferential stress on the ZP balloon material from the internal helium gas and the SP balloon inside. The arc length has three partitions:
\begin{enumerate}
    \item The base section $(s<s_0)$: Here the ZP balloon shape lies against the SP balloon envelope. Here the ZP circumferential stress is negative, compressing the two envelopes together. The initial meridional stress from the gondola weight and net buoyancy from the internal SP balloon is applied to this section (see Appendix \ref{app:BuoyancyForce}).
    \item The mid section $(s_0<s<s_l)$: Here the ZP balloon shape is loose and the circumferential stress in the ZP envelope is exactly or very nearly zero, as the envelope is in equilibrium.
    \item The apex section $(s>s_l)$:  Here the ZP envelope is taut. Here the ZP circumferential stress is positive, pushing the ZP balloon shape to its maximum extent. Physically, this happens because the helium gas inside is buoyant, and therefore collects at the top. As the ZP balloon envelope has high elastic modulus it does not stretch enough to meaningfully affect the shape, so when the balloon material is pulled taught it will very nearly follow the as-fabricated fully inflated geometry.
\end{enumerate}
Accordingly, for the  ``base section'' the FLOATS model can assume the ZP balloon lies against the known SP spherical geometry, and at the ``apex section'' the FLOATS model can assume the ZP balloon takes the known shape of the ZP fabricated geometry. The ``mid section shape'', and the location of its boundaries, must be computed. 

This mid section of the balloon shape is governed by the following set of ordinary differential equations in terms of the arc length $s$ for four variables - the shape angle, normalized meridional stress, radius from central axis, and height:
\begin{align} \label{eq:ode}
    \frac{\dd \theta_\text{ZP}}{\dd s} &=  -qrw_\text{ZP} \sin(\theta_\text{ZP})-qrb(z+z_{p0}) \nonumber\\
    \frac{\dd q}{\dd s} &= -q^2w_\text{ZP}r\cos(\theta_\text{ZP}) \nonumber\\
    \frac{\dd r}{\dd s} &= \sin(\theta_\text{ZP}) \nonumber\\
    \frac{\dd z}{\dd s} &= \cos(\theta_\text{ZP})
\end{align}
The point $z_{p0}$ a vertical offset to account for the height at which the internal and external pressures are exactly equal within the zero-pressure balloon. Secondary quantities $b$ and $q$ are defined as:
\begin{align}
    \rho_\text{diff} &= \rho_\text{atm} - \rho_\text{gas,ZP} \nonumber\\
    b &= g \rho_\text{diff} \nonumber\\
    q &= \frac{1}{r \sigma_m}
\end{align}
The ordinary differential equation Eq. \ref{eq:ode} is then subject to the following starting boundary conditions of the shape at $s=s_0$:
\begin{align}
\label{eq:bvp_start}
    s_0 &= R_\text{SP}\beta \nonumber \\
    \theta_\text{ZP}(s_0) &= \frac{\pi}{2}-\beta \nonumber \\
    q(s_0) &= \frac{2\pi \sin(\beta)}{F_\text{tension}(s_0)} \nonumber \\
    r(s_0) &= R_\text{SP}\sin(\beta) \nonumber \\
    z(s_0) &= R_\text{SP}(1-\cos(\beta))
\end{align}
and ending boundary conditions at $s=s_\ell$ where it hits the fully inflated geometry of the ZP balloon:
\begin{align}
\label{eq:bvp_end}
    \theta_\text{ZP}(s_\ell) &= \theta_\text{infl}(s_\ell) \nonumber \\
    r(s_\ell) &= r_\text{infl}(s_\ell)
\end{align}
The initial film tension $F_\text{tension}(s_0)$ is a critical parameter at the domain start, given by:  
\begin{align}
\label{eq:TAF}
    F_\text{tension}(s_0) &= g\Big(m_\text{payload}+w_\text{ZP} A_\text{ZP} (s_0)+w_\text{SP} A_\text{SP} + m_\text{SP,gas} - \frac{4}{3}\pi\rho_\text{ZP} R^3_\text{SP}\Big) - F_{\text{buoy}}(s_0) \nonumber \\
    F_{\text{buoy}}(s_0) &= \pi \sin^2(\beta) g\rho_\text{diff}\Big(z_{p0}-R_\text{SP}\Big)R_\text{SP}^2 +\frac{2\pi}{3}\Big(1-\cos^3(\beta )\Big)\rho_\text{diff}g R_\text{SP}^3 \nonumber \\
    A_\text{ZP} (s_0) &= 2 \pi R_\text{SP} z(s_0) \nonumber \\
    A_\text{SP} &= 4 \pi R^2_\text{SP}
\end{align}
where $F_{\text{buoy}}(s_0)$ is the net buoyancy force from the SP balloon and gondola payload acting on aerobot bottom surface from the base until $s_0$  that reduces the tension on the ZP material; the derivation for this force is given in the supplemental material. 

This set of differential equations and boundary conditions constitute a boundary value problem (BVP). In the above BVP, the values of $\rho_\text{diff}$ and $s_\ell$ are inputs. The values of $\beta$ and $z_{p0}$ are unknowns that are solved simultaneously with the BVP itself. The BVP for the balloon-in-balloon shape problem is a slightly more complex version of the BVP given in chapter four of \citet{LeakeDissertation}, where it was found that the Theory of Functional Connections (TFC) provides a speed improvement over a typical shooting method. For this problem, TFC was implemented numerically using the TFC Python module \cite{TfcGithub}. For readers interested in a mathematical introduction to TFC, see \citet{leake2022theory}. As solving this problem is computationally slower than other parts of the integrated FLOATS model, shapes are generated a-priori as a lookup table and then interpolated for flight simulations.

A fundamental limitation of this model is that it ignores the excess slack (i.e. ruffles of material) around the ZP balloon circumference, instead assuming the shape is perfectly axisymmetric with a gas bubble diameter that varies with height. While our model is therefore sufficient for the gross estimation of projected areas, and matches well with photo outlines of the balloon shape, these ruffles can add drag and asymmetries in film loading. We accordingly design the aerobot to higher wind and tensile load than expected in flight with extra layers of Kapton near the endfittings to spread the point load, and use field data to tune and validate the performance.


\subsubsection{Aerodynamic Model}
From the governing equation Eq. \ref{eq:eom}, the primary aerodynamic forces exerted on the system are the drag and virtual mass, expressed in detail by \citet{horne}. Notably, while these forces are significant for the balloon-in-a-balloon configuration, they are negligible for the payload given the vast difference in volume and drag area. 

The present model utilizes constant drag coefficients which are subject to fluctuations in the projected areas of the ZP balloon, denoted as $A_\text{top}$ and $A_\text{side}$ from the shape model in Sec. \ref{subsubsec:shape}. Virtual mass, caused by the momentum of the air accelerated out of the way during aerobot motions, is modeled as a constant coefficient $C_m$ which is then scaled by the varying volume of the aerobot. 

\begin{equation}
\label{aero}
F_\text{drag} = \frac{1}{2} \rho_\text{atm} C_D A_\text{ref} v^2 \hat{v}
\end{equation}

\begin{equation} \label{eq:cd}
\begin{aligned}
C_{D_\text{ref,top}} &= C_{D_\text{top}} \cdot \frac{A_\text{top}}{A_\text{ref}} \\
C_{D_\text{ref,side}} &= C_{D_\text{side}} \cdot \frac{A_\text{side}}{A_\text{ref}} \\
C_D &= \begin{bmatrix} C_{D_\text{ref,side}} & 0 & 0 \\ 0 & C_{D_\text{ref,side}} & 0 \\ 0 & 0 & C_{D_\text{ref,top}} \end{bmatrix}
\end{aligned}
\end{equation}

\begin{equation} \label{eq:ma}
\begin{aligned}
m_\text{virtual} = C_m \rho_\text{atm} V
\end{aligned}
\end{equation}

The aerobot velocity $v$ has coordinates in the balloon coordinate frame, wherein the $z$-axis points out the top, and the $x$- and $y$-axes point out the sides.

Choosing a $C_{D_\text{top}}$ and $C_{D_\text{side}}$ is difficult for balloons, as the flexible outer envelope generally does not support re-attachment of a turbulent boundary layer (unlike a sphere) \cite{horne}. Furthermore, the "Morison Equation" employed in Eq. \ref{eq:eom} for the drag and virtual mass is effective for steady-state forces and initial transient forces, but less so for the intermediate-time forces \cite{newman2018marine}. As such, the virtual mass and drag coefficients need to be tuned to experimental data at the appropriate movement timescale \cite{sarpkaya2010wave}. We choose $C_{D_\text{top}} = 0.8$ and $C_{D_\text{side}} = 1$ given good agreement with the indoor calibration flight data in \citet{izraelevitz2022subscale}, a $C_m = 0.2$ from ellipsoidal data in \citet{hoerner1958fluid}, and leave higher fidelity fluid-structure interaction in computation fluid dynamics (CFD) simulations to future work on how these coefficients should change independent of reference areas. 

\subsubsection{Thermodynamics Model}
\begin{figure}[t!]
\centering
\includegraphics[width=0.45\textwidth]{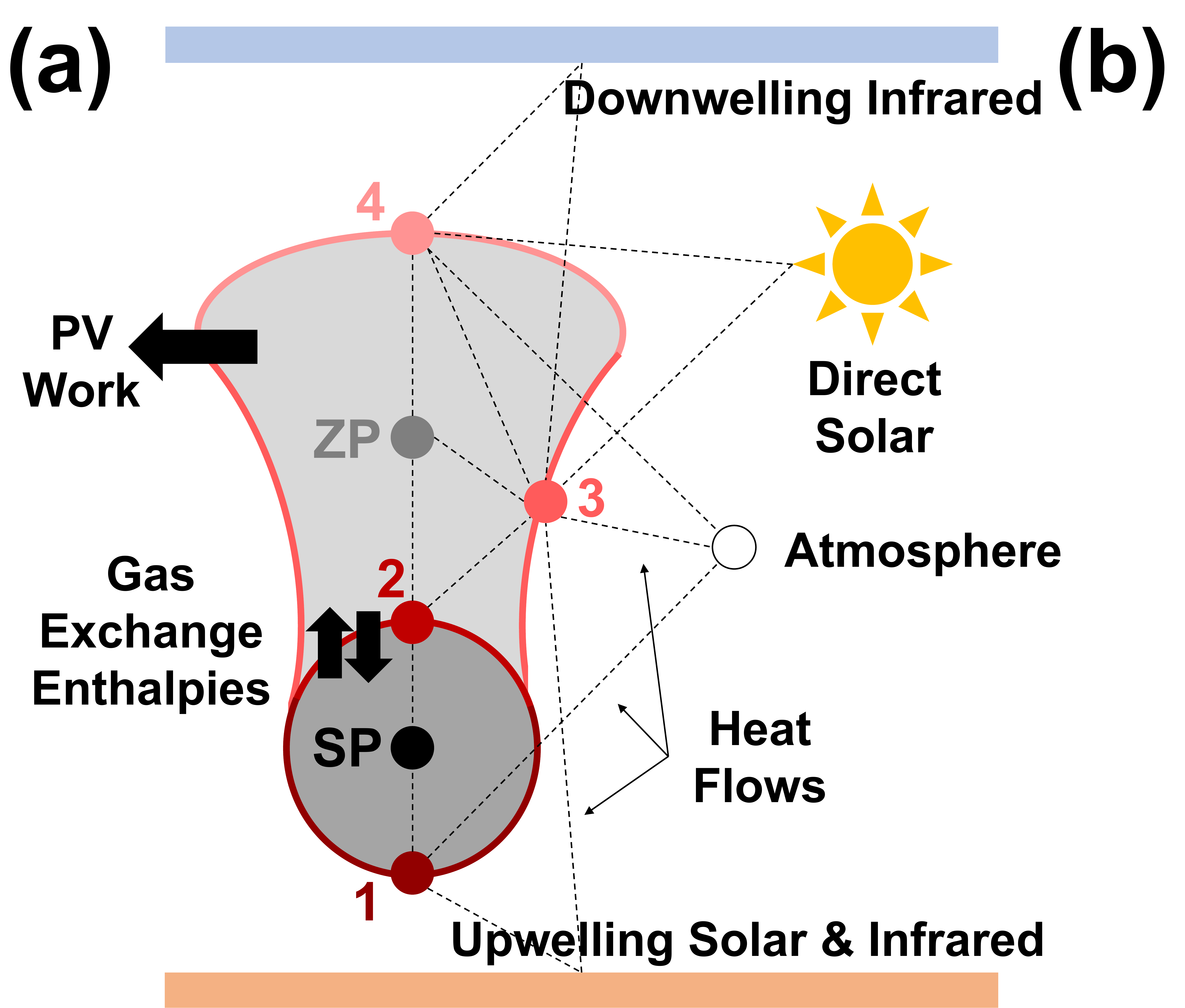}
\includegraphics[width=0.5\textwidth]{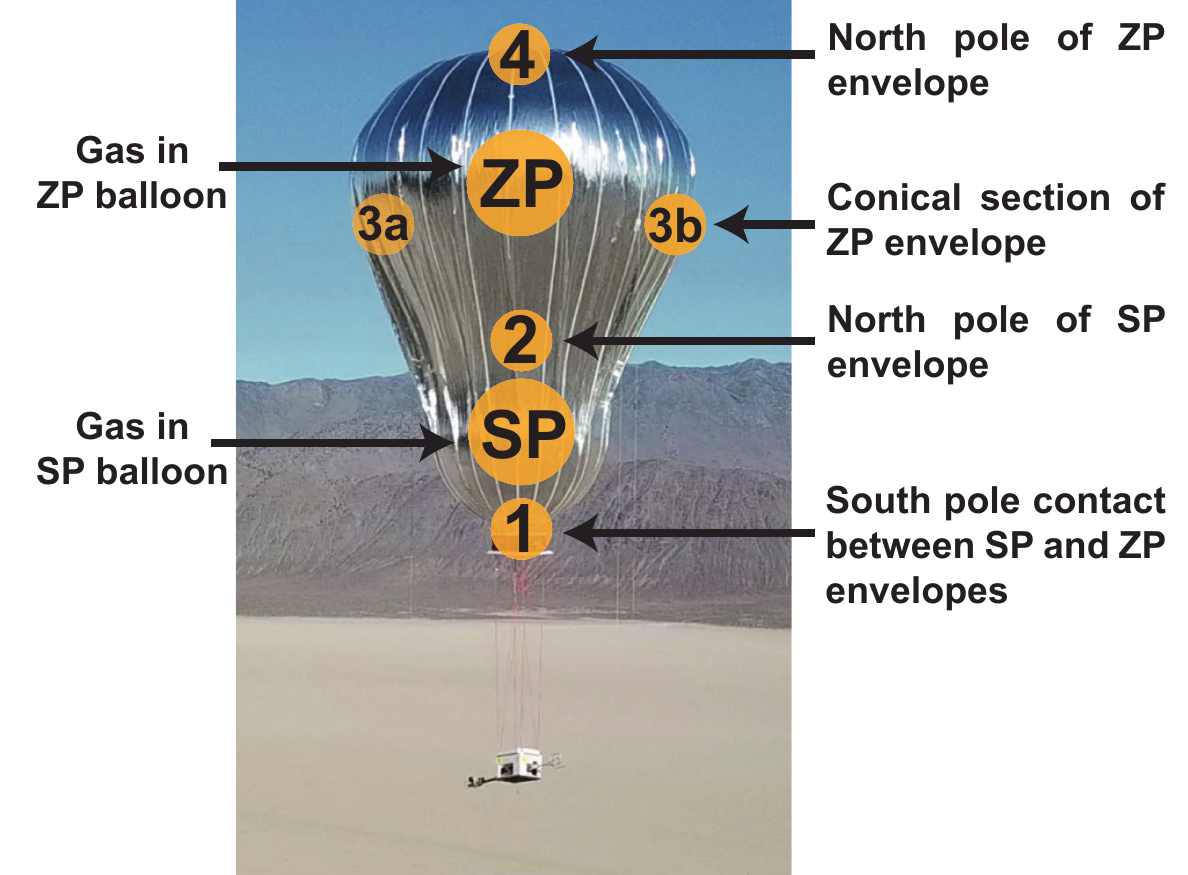}
\caption{Thermal nodes and control volumes used in thermodynamic modeling. (a) Work and gas enthalpies, along with the six thermal nodes ($i=1,2,3,4,$ ZP, SP). (b) Temperature sensor placement, with the exception of Node 2 and Node SP which were not measured due to access constraints. Node 3 was measured in two locations 3a and 3b on opposite sides. Photo courtesy NASA/JPL-Caltech, \copyright 2022 California Institute of Technology.}
\label{fig:biabnodes}
\end{figure}

The FLOATS thermodynamics model assumes that the temperatures for the envelopes and the gases are spatially averaged values, and that both the balloon and atmospheric gases are ideal. Figure \ref{fig:biabnodes} depicts the six thermal nodes and the following subsection will provide temperature rates-of-change of these nodes.

For the SP and ZP gas nodes, we start with a general control volume analysis of an open system (Fig. \ref{fig:biabnodes}a), including heat fluxes, work, and enthalpy flows of gas crossing in/out of the volume as:
\begin{equation}
    \label{eq:contrlVol}
    dU = \delta Q - \delta W + dH_\text{in} - dH_\text{out}
\end{equation}

Applying the control volume analysis to the SP balloon we first note that the SP balloon is approximately constant volume for the pressures experienced in flight and there is no shaft work, so $\delta W=0$. We therefore reach the following expression so FLOATS can integrate the SP temperature as:
\begin{equation}
\begin{gathered} \label{eq:dTsp}
\frac{d}{dt}U_\text{SP} = \frac{d}{dt}\left({m}_\text{SP,gas} c_v T_\text{SP}\right) = \dot{q}_\text{SP} + \Delta\dot{H}_\text{SP} \\
\boxed{\dot{T}_\text{SP} = \frac{-\dot{m}_\text{SP,gas}  c_v  T_\text{SP} + \dot{q}_\text{SP} + \Delta\dot{H}_\text{SP} }{ m_\text{SP,gas} c_v }}
\end{gathered}
\end{equation}
The net enthalpy rate $ \Delta\dot{H}_\text{SP}$ for the gas crossing the SP balloon boundary is derived from the vent's input and pump's output enthalpies:
\begin{equation}
\label{eq:Hsp}
\Delta\dot{H}_\text{SP} = \dot{H}_\text{pump,out} - \dot{H}_\text{vent,in}
\end{equation}

Moving to the ZP balloon, the control volume analysis of this system now includes the boundary work $\delta W=PdV$ of the expanding balloon, giving the following expression for FLOATS to integrate the ZP temperature as:
\begin{equation}
\label{eq:dTzp}
\boxed{\dot{T}_\text{ZP} = \frac{-\dot{m}_\text{ZP,gas}  c_v  T_\text{ZP} + \dot{q}_\text{ZP} - P_\text{ZP} \dot{V}_\text{ZP} + \Delta\dot{H}_\text{ZP} }{m_\text{ZP,gas} c_v }}
\end{equation}
The enthalpy rates for the gas crossing the ZP balloon boundary, $\Delta \dot{H}_\text{ZP}$ is specified as:
\begin{equation}
\label{eq:Hzp}
\Delta\dot{H}_\text{ZP} =  \dot{H}_\text{vent,out} - \dot{H}_\text{pump,in}
\end{equation}

The enthalpy flows through the pump and vent are a function of the mass flow rates and temperatures on each end. Typically compression pumps include cooling fins to increase efficiency, so we assume an adiabatic pump process to overpredict (i.e. conservatively estimate) the temperature increase $T_\text{pump,out}$ as: 
\begin{align}
\label{eq:Tp}
\dot{H}_\text{pump,in} &= \dot{m}_\text{pump}  c_p  T_\text{ZP} \nonumber \\
\dot{H}_\text{pump,out} &= \dot{m}_\text{pump}  c_p  T_\text{pump,out} \nonumber \\
T_\text{pump,out} &= T_\text{ZP} \cdot \left( \frac{P_\text{SP,abs}}{P_\text{atm}} \right)^{\frac{k-1}{k}}
\end{align}
Note the use of $c_p$ for enthalpies $H$, $c_v$ for internal energies $U$, and $k=c_p/c_v$. The vent between balloons we assume to be isenthalpic, as is usual for the free venting of a pressure vessel:
\begin{align}
\label{eq:Hv}
\dot{H}_\text{vent,in} &= \dot{m}_\text{vent}  c_p  T_\text{SP} \nonumber \\
\dot{H}_\text{vent,out} &= \dot{H}_\text{vent,in}
\end{align}
Some simplification of the gas temperature in Eqs. \ref{eq:dTzp} and \ref{eq:dTsp} is possible using $c_p = c_v + \overline{R}$ as in other balloon thermodynamics derivations \cite{horne,kreith1974numerical}, but keeping everything explicit allows for full enumeration of the enthalpy paths. 

Finally, for the envelope material temperature on the boundaries of the SP and ZP balloons, the temperature rates can be generalized to a single equation that can apply to all four envelope nodes $i=1,2,3,4$ as:
\begin{equation}
\label{eq:Tnodes}
\boxed{\dot{T}_i = \frac{\dot{q}_i}{m_i c_i}}
\end{equation}

\subsubsection{Heat Transfer Model}
The heat transfer model considers convection and radiation for all nodes (Fig. \ref{fig:biabnodes}a). Accordingly, the lifting gas nodes (SP \& ZP) see convective and radiative heating to the envelope, and the envelope nodes (i = $1, 2, 3, 4$) see convective and radiative heating to both the lifting gas and the outside environment. Starting with the gas nodes, their total heat flux is given as:
\begin{align}
\label{eq:qspzp}
\dot{q}_\text{ZP} &= \sum_{i=2,3,4} \left( \dot{q}_{\text{infrared, } i \rightarrow \text{ZP}} +  \dot{q}_{\text{convection,  }i \rightarrow \text{ZP}} \right)  \nonumber \\
\dot{q}_\text{SP} &= \sum_{i=1,2} \left( \dot{q}_{\text{infrared, } i \rightarrow \text{SP}} +  \dot{q}_{\text{convection,  }i \rightarrow \text{SP}} \right)  
\end{align}

Note that these two gas nodes (SP \& ZP) do not see the outside environmental fluxes directly. The heat flows to each of the four envelope nodes on their boundaries ($i=1,2,3,4$) are then given by convective and radiative (both solar and infrared) terms to the outside environment:
\begin{align}
\label{eq:qi}
\dot{q}_\text{1} &= \dot{q}_{\text{solar, 1, upwelling}} + \left(\dot{q}_{\text{infrared, 1, upwelling}} - \dot{q}_{\text{infrared, }1, \text{out} }\right) + \sum_{i=\text{atm},\text{SP}}\dot{q}_{\text{convection, } i\rightarrow 1} +  \sum_{i=2,\text{SP}} \dot{q}_{\text{infrared, } i\rightarrow 1 } \nonumber \\
\dot{q}_\text{2} &= \sum_{i=\text{SP, ZP}}\dot{q}_{\text{convection, } i\rightarrow 2} +  \sum_{i=1,3,4,\text{SP, ZP}} \dot{q}_{\text{infrared, } i\rightarrow 2 } \nonumber \\
\dot{q}_\text{3} &= \dot{q}_{\text{solar, 3, sidewelling}} + \left(\dot{q}_{\text{infrared, 3, sidewelling}} - \dot{q}_{\text{infrared, }3, \text{out} }\right) + \sum_{i=\text{atm, ZP}}\dot{q}_{\text{convection, } i\rightarrow 3} +  \sum_{i=2,4,\text{ZP}} \dot{q}_{\text{infrared, } i\rightarrow 3 } \nonumber \\
\dot{q}_\text{4} &= \dot{q}_{\text{solar, 4, downwelling}} + \left(\dot{q}_{\text{infrared, 4, downwelling}} - \dot{q}_{\text{infrared, }4, \text{out} }\right) + \sum_{i=\text{atm, ZP}}\dot{q}_{\text{convection, } i\rightarrow 4} +  \sum_{i=2,3,\text{ZP}} \dot{q}_{\text{infrared, } i\rightarrow 4 } 
\end{align}
The fluxes impinging on the external envelope of the aerobot are broken into upwelling, downwelling, and direct irradiances $E$ of the environment from three sources (illustrated in Fig. \ref{fig:biabnodes}): 
\begin{enumerate}
    \item \textit{A planar ground source} in both upwelling longwave $E_\text{upwelling, infrared}$ (ground temperature) and upwelling shortwave  $E_\text{upwelling, solar}$ (ground albedo).
    \item \textit{A planar sky source} in downwelling longwave $E_\text{downwelling, infrared}$ (sky temperature).
    \item \textit{A solar point source} in direct shortwave $E_\text{direct, solar}$ (sunlight). 
\end{enumerate}

Each flux source is then paired with specific view factors for each node. External solar viewfactors assume the balloon approximates a circular cylinder flying over a flat diffuse ground source and a point solar source, where upwards and downwards facing surfaces see full area illumination but only 1/$\pi$ of the side area is hit by the direct sun: 
\begin{align}
\label{eq:qsolar}
\dot{q}_{\text{solar, 1, upwelling}} &= \alpha A_1 E_\text{upwelling, solar} \nonumber \\
\dot{q}_{\text{solar, 3, sidewelling}} &= \alpha A_3 \left(\frac{1}{2}E_\text{upwelling, solar} +  \frac{1}{\pi}E_\text{direct, solar} \sin{\theta_\text{sun}}\right) \nonumber \\
\dot{q}_{\text{solar, 4, downwelling}} &= \alpha A_4 \left( E_\text{direct, solar} \cos{\theta_\text{sun}}\right)
\end{align}
The infrared viewfactors assume diffuse upwards and downwards sky and ground sources as:
\begin{align}
\label{eq:qir}
\dot{q}_{\text{infrared, 1, upwelling}} &= \epsilon A_1 E_\text{upwelling, infrared} \nonumber \\
\dot{q}_{\text{infrared, 3, sidewelling}} &= \epsilon A_3 \left(\frac{1}{2} E_\text{upwelling, infrared} + \frac{1}{2} E_\text{downwelling, infrared}\right) \nonumber \\
\dot{q}_{\text{infrared, 4, downwelling}} &= \epsilon A_4 E_\text{downwelling, infrared} \nonumber \\
\dot{q}_{\text{infrared, i, out}} &= \epsilon_i A_i \sigma T^4_i
\end{align}
while directional infrared fluxes between nodes on the aerobot interior are given by the radiative heating relation of \cite{lienhard2024heat}:

\begin{equation}
\label{eq:qIR}
\dot{q}_{\text{infrared, } i\rightarrow j } =  \sigma \left(T^4_i - T^4_j \right)\left/ \left(\frac{1-\epsilon_i}{\epsilon_i A_i} + \frac{1}{f_{i\rightarrow j} A_i} + \frac{1-\epsilon_j}{\epsilon_j A_j}\right) \right.
\end{equation}

For the convection terms, the convective heating between an envelope node and a gas node assumes a Nusselt formulation for a sphere. The natural convection Nusselt relation is chosen for ZP and SP helium gases to their boundary envelopes, and forced convection for transfer to the atmosphere:
\begin{align}
\label{eq:convection}
\dot{q}_{\text{convection, } \text{ZP} \rightarrow j } &=  \frac{\text{Nu}_\text{natural} \cdot k_\text{gas}} {l_{\text{ZP} \rightarrow j}} A_{\text{ZP} \rightarrow j} (T_\text{ZP}-T_j) \nonumber \\
\dot{q}_{\text{convection, } \text{SP} \rightarrow j } &=  \frac{\text{Nu}_\text{natural} \cdot k_\text{gas}} {l_{\text{SP} \rightarrow j}} A_{\text{SP} \rightarrow j} (T_\text{SP}-T_j) \nonumber \\
\dot{q}_{\text{convection, atm } \rightarrow j } &=  \frac{\text{Nu}_\text{forced} \cdot k_\text{atm}} {l_{\text{atm} \rightarrow j}} A_{\text{atm} \rightarrow j} (T_i-T_j) \nonumber \\
\text{Nu}_\text{natural} &= 2.0 + 0.6 \text{Ra}^{0.25} \nonumber \\
\text{Nu}_\text{forced}  &= \left\{
    \begin{matrix}
        0.37 \text{Re}^{0.6} &\text{Re}\leq5.38 \cdot 10^5 \\
        0.74 \text{Re}^{0.6} &\text{otherwise}
    \end{matrix} \right\}
\end{align}

\subsubsection{Atmospheric model \label{ssec:atmosmodel}}
For pre-flight simulations, the EarthGram atmospheric model was utilized to predict the environmental conditions \cite{white2023earth}. Developed by the National Center for Atmospheric Research (NCAR), the EarthGram model is a detailed global representation of the Earth's topography and atmosphere -- encompassing its composition, structure, and dynamics - and it is widely utilized in weather and climate research. FLOATS altitude predictions in EarthGram, along with lateral winds from NOAA GFS 0.25 degree, were used for the pre-flight planning in Section \ref{subsec:planning}.

\begin{figure} [t]
    \centering
    \includegraphics[width=0.95\textwidth]{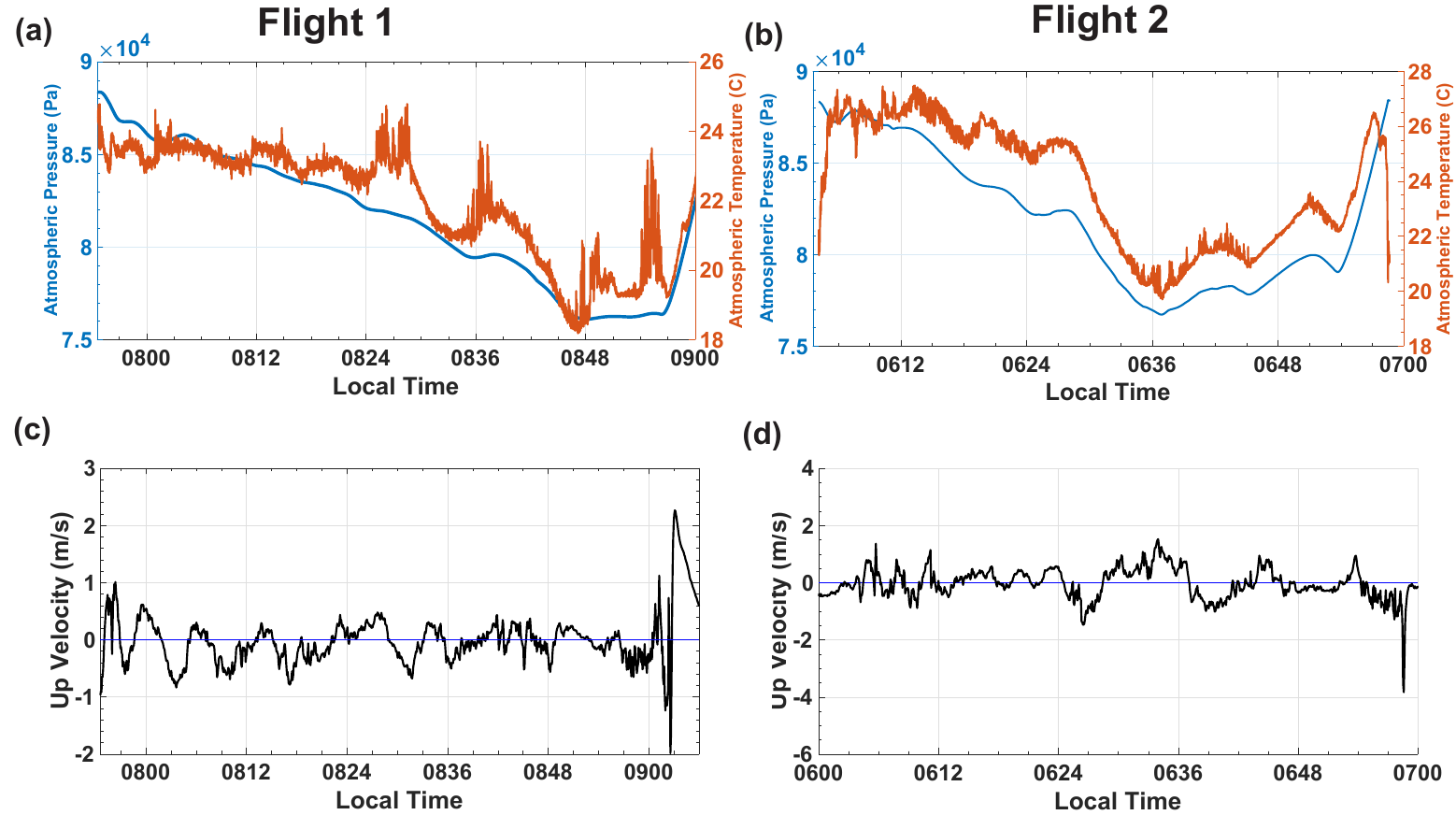}
    \caption{Atmospheric pressure, temperature, and inertial-frame winds recorded from (a, c) Flight 1 and (b, d) Flight 2. Wind velocity in the inertial frame were computed by vectorially adding the vertical velocity of the balloon with the wind velocity recorded on the anemometer on the gondola.}
    \label{fig:atmosphere}
\end{figure}

\begin{figure} [t]
    \centering
    \includegraphics[width=\textwidth]{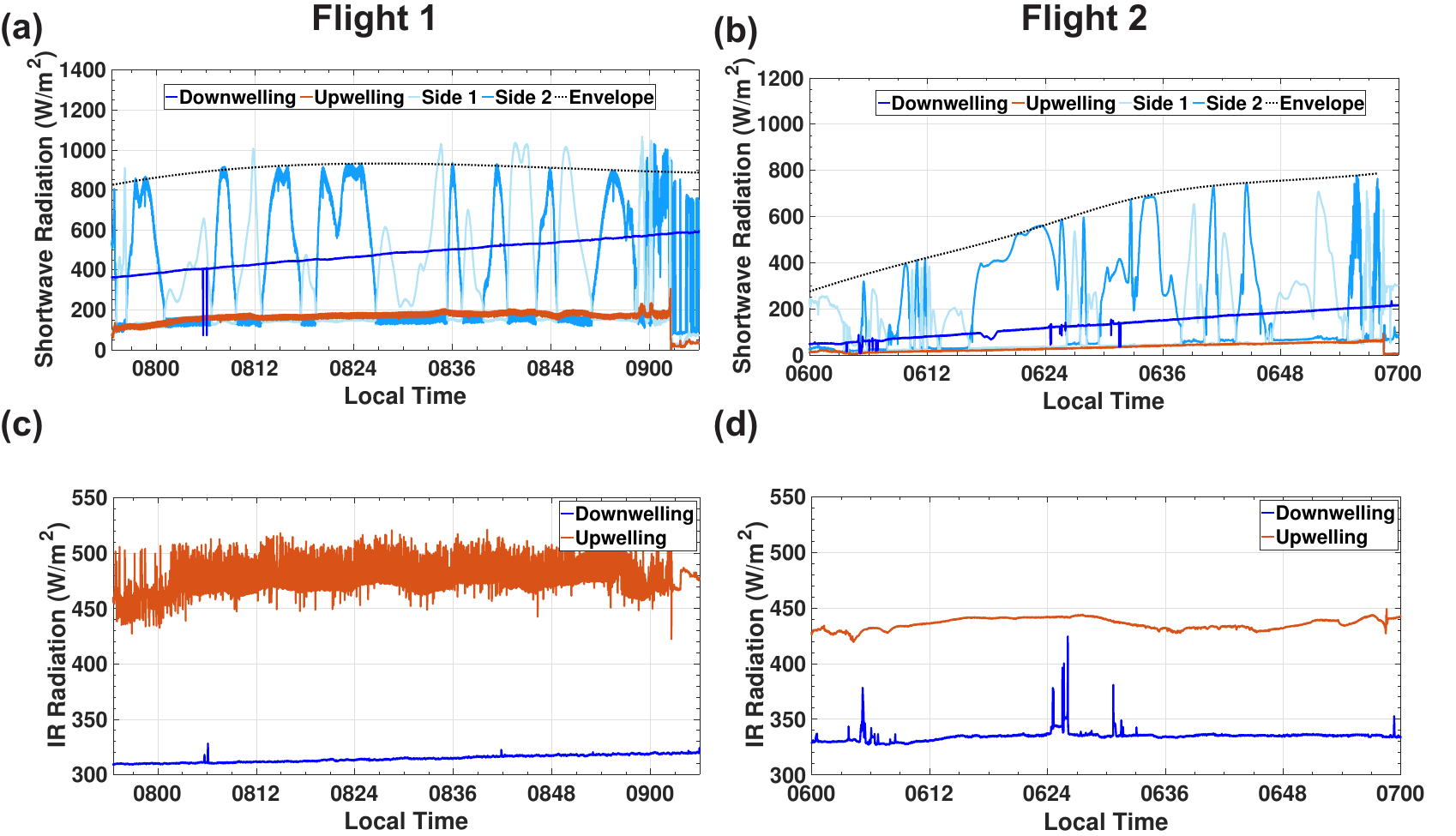}
    \caption{Solar radiation from pyranometers for (a) Flight 1 and (b) Flight 2. Longwave IR from pyrgeometers for (c) Flight 1 and (d) Flight 2. Upwelling and side radiation are measured on the gondola (150$^\circ$ and 180$^\circ$ FOV respectively), downwelling radiation from the ground station with 180$^\circ$ FOV. }
    \label{fig:radiation}
\end{figure}

In post-flight analysis, the multi-sensor dataset collected in the outdoor free flights was used to overwrite EarthGram as new inputs to the FLOATS model to account for the exact weather of the day. Figures \ref{fig:atmosphere} and \ref{fig:radiation} show as-measured atmospheric property inputs such as temperature, pressure, winds, and solar radiation supplied to the FLOATS model. Data used within the FLOATS model were corrected based on post-flight calibration as discussed in Appendix \ref{sec.Appendix}. Venting and pumping times were supplied to the FLOATS model based on data collected on board as well. 

Atmospheric temperature was measured via two methods: (1) an ultrasonic anemometer sonic temperature and (2) a temperature sensor exposed to the surrounding air. The temperature sensor was observed to have a large response time lag (rise time of approximately 4 minutes leading to errors of up to $5^\circ$C) and, in comparison, the sonic temperature showed more consistent values between ascent and descent albeit some noise. Accordingly, a filtered spline fit of the sonic temperature (Figure \ref{fig:atmosphere}a-b, red) was used as temperature input to FLOATS.

The ultrasonic anemometer also provided 3-D winds measured on the aerobot gondola (Figure \ref{fig:atmosphere}c-d, black). These were rotated into the inertial frame and vectorially added to the aerobot inertial velocity (obtained from the INS on the gondola) to generate inertial wind velocities, which were then used by FLOATS. Ground winds from the cup anemometer at the launch site were not logged, only used for launch go/no-go decisions.

Diffuse shortwave and longwave radiation were measured from a variety of locations (Figure \ref{fig:radiation}) and then assigned to irradiance sources for FLOATS. The upwelling longwave and upwelling shortwave (ground temperature \& albedo) was determined from the pyranometers and pyrgeometers on the gondola. The downwelling longwave (sky temperature) was determined from the ground station pyrgeometer. Accidental shadows cast on the groundstation by the field team were filtered out in post-processing.  For both our flights, this sky temperature was cooler than the ground temperature.

Lastly, the direct shortwave (sunlight) was computed from a combination of all four pyranometers as none permanently face the morning sun, and the Lambert cosine effect on all sensors must accordingly be inverted. A sidewelling shortwave flux magnitude $E_\text{sidewelling, solar}$ was first computed as an envelope function from the side-mounted pyranometers, accounting for the motion in and out of the sunlight as the aerobot rotates. Subsequently, the upwelling flux from the bright ground was subtracted from this measured sideways signal with a viewfactor of 1/2 to provide a sun-only horizontal flux, and then direct solar magnitude was derived by removing the solar angle effect as:
    \begin{align}
        \label{eq:directsolar}
E_{\text{direct, solar}} &= \sqrt{\left(E_{\text{direct, solar}}\sin{\theta_\text{sun}}\right)^2+\left(E_{\text{direct, solar}}\cos{\theta_\text{sun}}\right)^2} \nonumber \\ &= \sqrt{\left(E_{\text{sidewelling, solar}} - \frac{1}{2} E_{\text{upwelling, solar}}\right)^2+E_{\text{downwelling, solar}}^2}
    \end{align}
    Given the early morning test flights with the sun close to the horizon and dim ground, however, the Lambert cosine effect has little influence. The resultant direct solar flux is nearly identical to measured sidewelling signal, but the distinction would be important for future longer duration flights closer to local noon.

All inputs could be supplied to FLOATS as functions of either altitude or time, depending on the phenomenon. We chose to parameterize the measured shortwave radiation and winds by time, as solar incidence angle is primarily time-dependent at low altitude, and the vertical wind gusts proved to be non-recurring disturbances. We decided to parameterize the measured atmospheric pressure and temperature by altitude instead of time (with high confidence for pressure and medium confidence on temperature; see Results Section \ref{sec:results}), as these properties showed repetition on the aerobot ascent and descent - presumably given the long timescale of the atmosphere change in the morning compared to the balloon motion. Upwelling longwave radiation was nearly constant, but showed minor recurrence between ascent and descent and therefore was also parameterized by altitude. Downwelling longwave was measured at the groundstation so therefore parameterized by time.

\section{Results \label{sec:results}}
\begin{figure} [t]
    \centering
    \includegraphics[width=\textwidth]{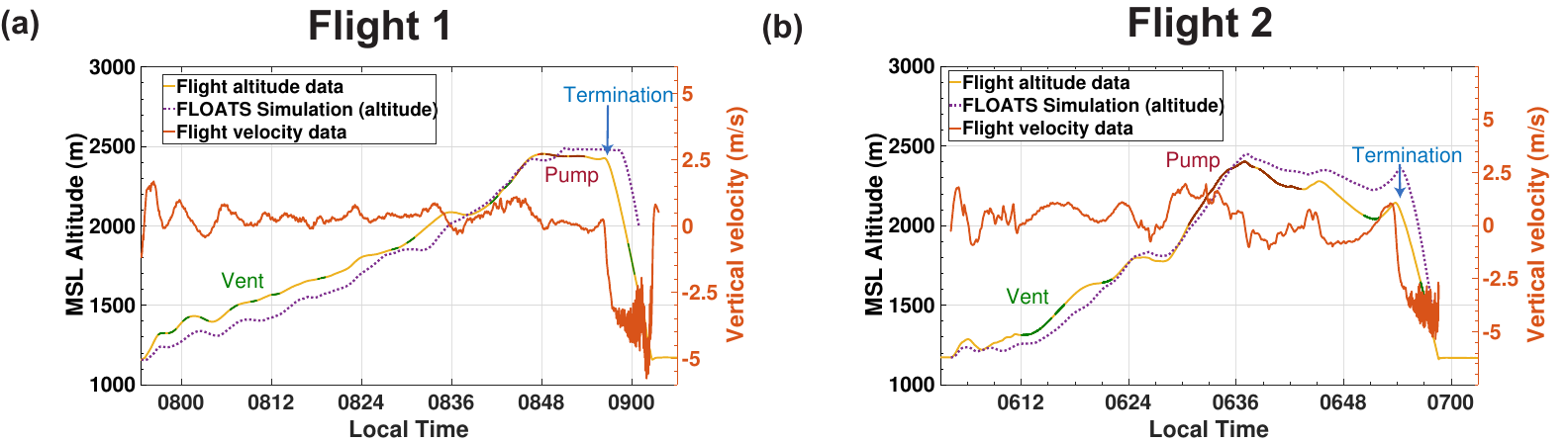}
    \caption{Simulated and observed altitudes and observed vertical velocity on (a) Flight 1 and (b) Flight 2. Green sections indicate venting, red sections indicate pumping, and blue indicates termination by the poppet. We assume a tailored 100 grams (Flight 1) and 420 grams (Flight 2) of free lift in FLOATS rather than measured values of 350 grams and 320 grams.}
    \label{fig:altitudes}
    \vspace{3mm}
    \includegraphics[width=\textwidth]{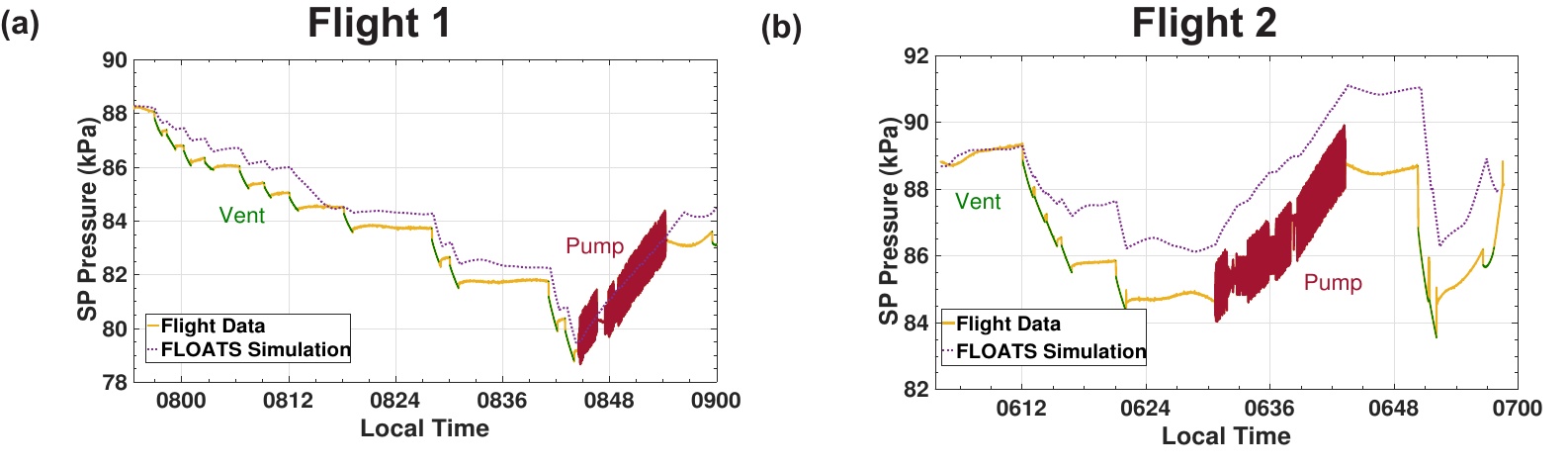}
    \caption{Simulated and observed SP balloon absolute pressure (not gauge pressure) for (a) Flight 1 and (b) Flight 2. As before, vent and pump actions are labeled in green and red respectively. Pump noise is an artifact of compression waves near the pressure transducer from the oscillating pump pistons. }
    \label{fig:superpressure}
\end{figure}

\begin{figure} [ht!]
    \centering
    \includegraphics[width=0.75\textwidth]{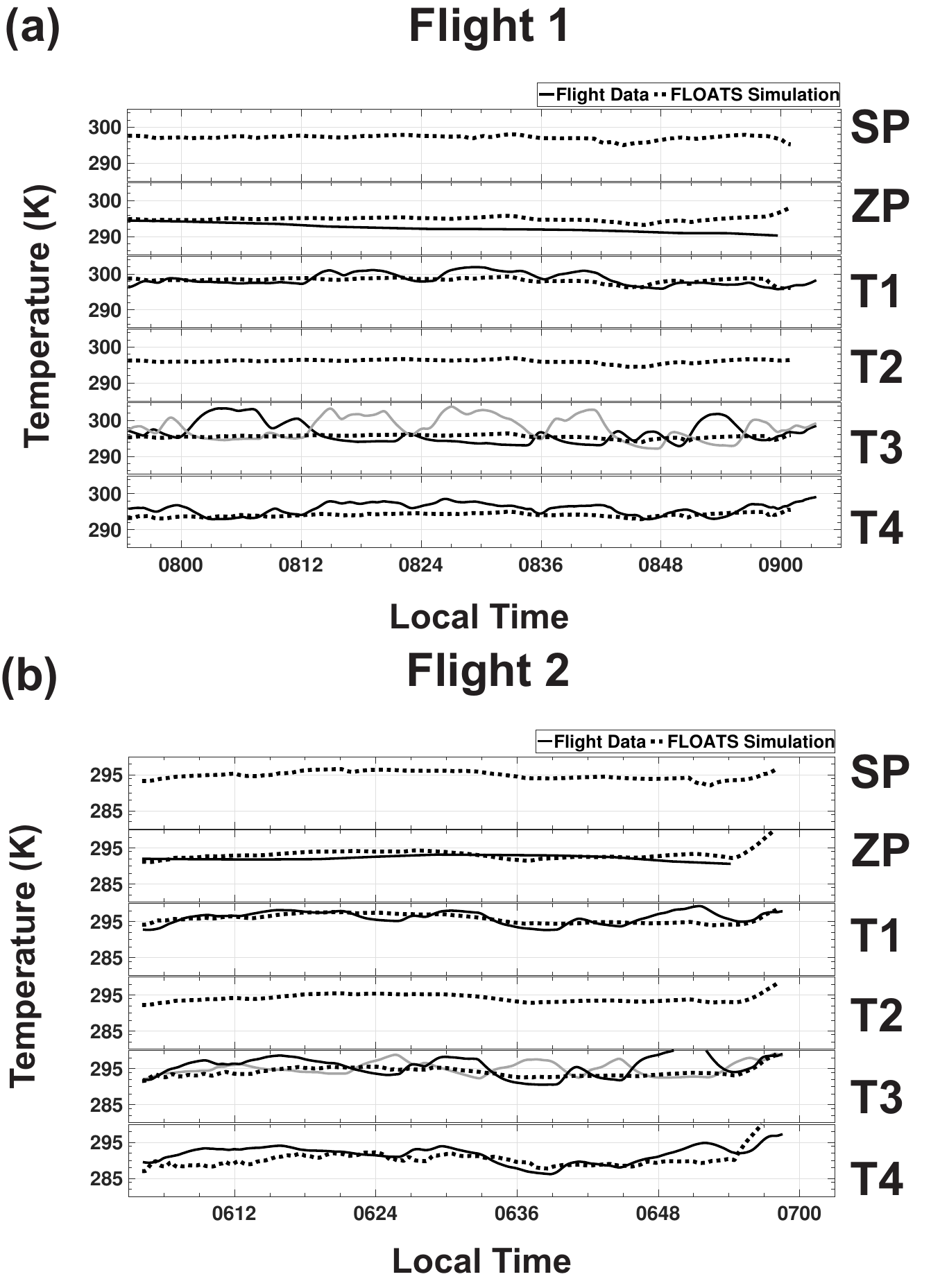}
    \caption{Simulated and observed node temperatures. Note that the T2 node and SP node were not measured, and T3 was measured in two locations on opposite sides of the ZP envelope (shown by black and gray solid lines). Corrections from the post-flight calibration were applied (see appendix)}
    \label{fig:temperature}
\end{figure}

The main output comparisons for performance evaluation of the FLOATS model (namely aerobot altitude, SP balloon pressure, and gas and skin temperature) are shown in Figures \ref{fig:altitudes}, \ref{fig:superpressure}, and \ref{fig:temperature}. Note that all plots zoom the Y-scale into the signal range in question to magnify the readability of data, and accordingly the zero value is not along the X-axis.

Aerobot altitude predictions from FLOATS (Fig.\ref{fig:altitudes}) have both been tailored with free-lift values (in increments of 50~g) to best match the maximum altitude from the FLOATS simulation, as uncertainties of order 300~g were possible given our experience with the measurement technique in the groundwinds.  For Flight 1, the over 4~m/s ground winds severely impacted our ability to accurately measure free lift, and tailored value of 100~g was significantly less than measured (350~g). In Flight 2 however, lower groundwinds meant our tailored value of 420~g was much closer to the measured value (320~g). We accordingly have much more confidence in altitude match of FLOATS in Flight 2. The phasing of rises and descents over the course of the flight matches well to the FLOATS simulation, meaning the instrumentation mostly captured the important short duration relevant inputs (winds/BCM actions etc). Accordingly, we believe the internal gas pressures were well measured and that the wind sensor was not shadowed by aerobot wake.

Fig. \ref{fig:superpressure} plots predicted and observed SP absolute pressure values in both flights. We show the absolute pressure value of the gas within the SP balloon (absolute = gauge pressure $+$ atmospheric pressure) as the gauge pressure is less diagnostic as it is sensitive to the precise altitude. We notice good agreement between the observed and simulated values, with approximately 2~kPa accumulated error over time ($\sim$ 2.5 $\%$). This error accumulation is dominated by the venting actions - the gas flow due to venting action is ZP pressure dependent (and therefore altitude dependent), so as FLOATS underestimates in altitude early in both flights it then underpredicts the vent actions. The SP gauge pressure rise rate due to the pump is captured better by the model, as it is a positive displacement pump with low sensitivity to altitude error. 

We also note good agreement (Fig. \ref{fig:temperature}) between the simulated and observed gas and skin temperatures with the ZP gas temperature. The temperature sensor measurements at the top and bottom of the aerobot are the most consistent with FLOATS predictions, whereas internal and side temperatures are less consistent. Typical thermal error between FLOATS predictions and temperature measurements is within $\pm 7^\circ$C, with better agreement during Flight 2 compared to Flight 1, and the most error is in the T3 sensors that cycle in and out of shadow rather than average into a single FLOATS node. The sudden temperature rise in FLOATS after activating the termination poppet is likely physical, as the ZP gas remaining after termination will adiabatically compress during rapid descent, but is not captured by the slow internal temperature logger that only logs every 5 minutes. In contrast, skin temperature sensors are sampled at 20 sps and averaged over 10 seconds to produce a time series at 0.1 sps do capture the rise. Future flights will emphasize better thermal access, especially within the ZP balloon to capture the thermally-driven buoyancy dynamics. 


In our analysis, we found that the heat transfer model within FLOATS is the largest source of sensitivity in the outputs and likely drives our modeling error. Notably, the simulated aerobot altitude showcases the strongest sensitivity to the atmospheric temperature in flight. FLOATS cannot currently accommodate a time-carying atmosphere and only accepts the measured atmospheric temperature with altitude -- thus variations in temperature at the same altitude during the approximately hour-long flights of the aerobot were averaged. This is a key limitation of the model, especially in the late morning as hot air near the ground begins to rise and mix, decreasing the air density. By contrast, FLOATS does simulate the rise in temperature of the ZP helium gas in sunlight. Accordingly, while FLOATS includes the increasing buoyancy effect from the warm helium it simultaneously ignores the decreasing buoyancy effect from the warm air.

The atmosphere was under weaker solar forcing in Flight 2 given a dimly lit earlier morning launch than Flight 1, so part of the FLOATS discrepancy could be explained by stronger time-varying atmosphere of Flight 1 that accentuates this fundamental limitation in FLOATS. The flight instrumentation measured this time-varying temperature phenomena on multiple occasions, with large atmospheric temperature excursions independent of altitude (Figure \ref{fig:atmosphere}a, red) whereas Flight 2 was much better altitude-correlated. We further observed during data reduction that the FLOATS altitude profile is markedly changed if the temperature-altitude spline fit varies by even 2--3\textsuperscript{$\circ$}C near the surface and would overshoot in altitude if the high-altitude thermal excursions are not captured in the averaged atmosphere. This type of model sensitivity to thermal conditions was not observed in the actual flight behavior of the aerobot, which flew as commanded by the pilot, so understanding this effect is our highest priority for further investigation. 

Future FLOATS modeling work will focus on these thermal excursions and requires longer duration flights for validation under many hours of variable solar forcing and atmospheric temperature change. Better capturing of the thermal time constant between the ZP gas and outside atmosphere (convective heat transfer rates), azimuthal nodes, and a time-varying atmospheric temperature model should improve the FLOATS prediction on how these variations impact the flight dynamics. Longer duration flights will also decrease the effect of free-lift measurement error, as uncertainty in the initial flight phase would get averaged over many hours.

\section{Discussion: Application to Venus \label{sec:discussion}}
The primary utility of FLOATS after validation against Earth flight test data is to predict flight trajectories on Venus, allowing simulation of a full-scale aerobot to aid mission planning studies. On Venus, the altitude control capability would be used by the science operations team to interrogate multiple altitudes over different times of day, allowing for a extensive characterization of the cloud layer with the onboard instrumentation. For supporting Venus simulations, the Venus Climate Database (VCD) \cite{lebonnois2016wave} has been integrated into FLOATS. The VCD is a comprehensive compilation of atmospheric data and models for Venus, encompassing information on the planet's atmospheric composition, structure, and dynamics, as well as data on surface features and climate patterns. Solar and infrared fluxes are taken from \citet{robinson2018linearized} as a function of local solar zenith angle and applied purely as diffuse sources, as little direct sunlight reaches the target altitudes.

A variety of full-scale Venus aerobot designs are described in \citet{hall2021prototype} with a 100~kg gondola and an altitude range from 52~km-62~km. These aerobots are approximately three times larger diameter than the subscale prototype flown in the Nevada experiment, and accordingly can be designed to cover the full altitude range even with this larger payload given their 27-fold increase in volume.

\begin{figure} [t]
    \centering
    \includegraphics[width=0.9\textwidth]{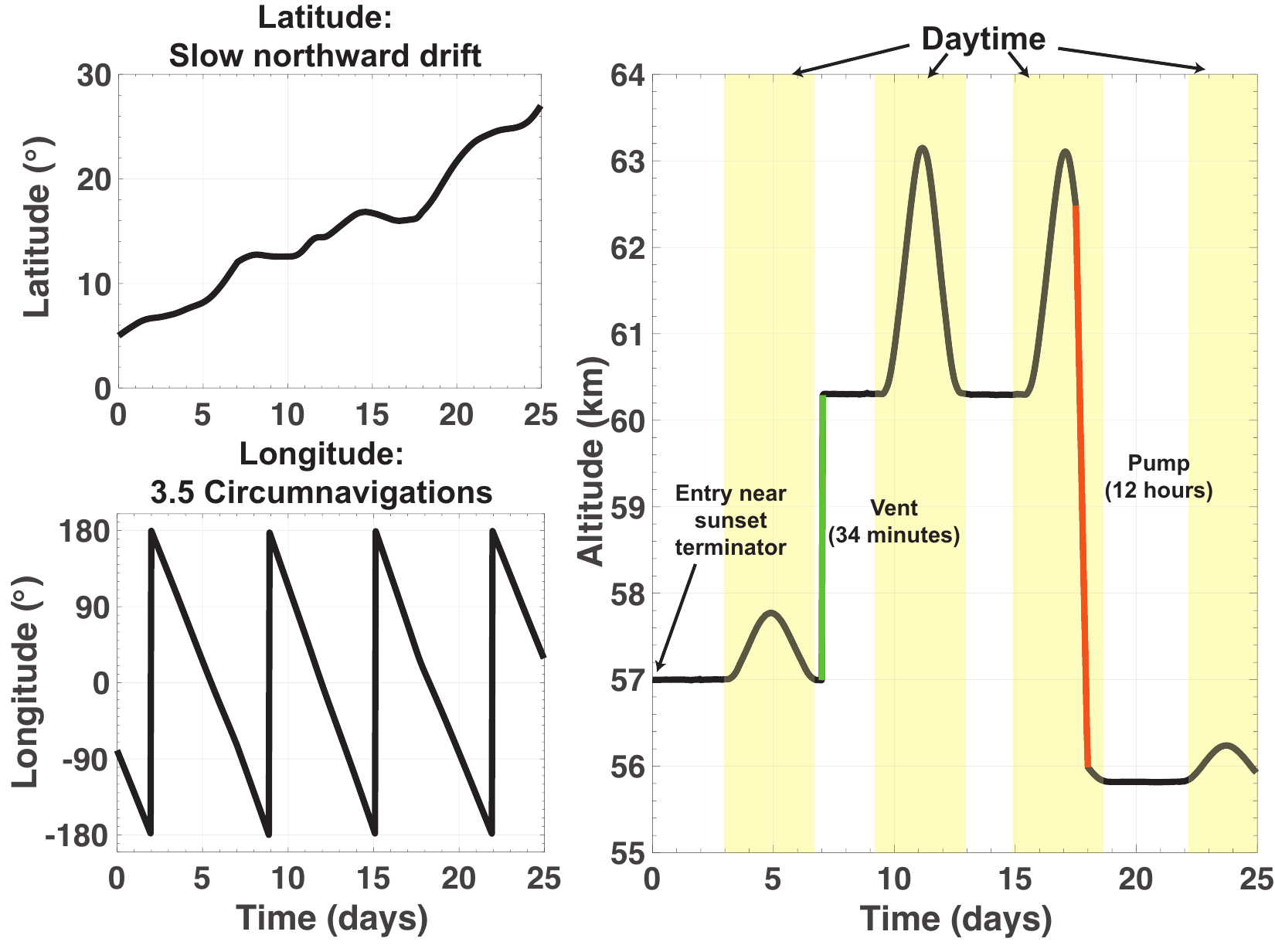}
    \caption{Simulated trajectory for a 12~meter diameter aerobot on Venus (100~kg payload capacity) for 25 Earth days. Left: Aerobot latitude and longitude. Right: Altitude over the Venus surface given a venting action after the first circumnavigation  and a pumping action after the third local noon. }
    \label{fig:venusflight}
\end{figure}

Current estimates from envelope coupon helium retention measurements are consistent with a 1-Venus-day flight duration (117 Earth days), but minimum science baseline would be closer to a few weeks. One such baseline Venus flight of a 12.5 meter diameter aerobot (taken from \citet{hall2021prototype} case B2) is simulated in Figure \ref{fig:venusflight}, illustrating three circumnavigations of the planet after entry at $5^\circ$ latitude at local sunset with zonal and meridional winds imported from VCD. 

As a caveat, extrapolation of a model from a 1~km flight test to a 10~km extent Venus performance is debatable even at similar atmospheric densities, so future flights with larger aerobot prototypes and a wider range of altitudes from Table \ref{tab:atmos} will be needed to cover this gap. At this development point, FLOATS simulations can only be an early hint at what performance should be. Large Monte Carlo simulations are supported in this codebase and are planned for future work, but this first flight example shows a number of salient features worthy of discussion. 

First, the passive altitude dynamics of the aerobot for global scale Venus winds and fluxes appear stable, especially at night. Solar heating during the day however does create an excess in ZP buoyancy, as the balloon material still retains a non-negligible $\alpha/\epsilon$ ratio. The equilibrium altitude accordingly tracks a kilometer-scale excursion as a function of the solar zenith angle peaking at local noon, which could be mitigated if desired through closed-loop altitude control in the BCM.    

Secondly, the venting and pumping gas between the SP and ZP balloons at specific chosen interval within the mission allows operators to move the equilibrium altitude. The flight in Figure \ref{fig:venusflight} includes one circumnavigation of the planet purely passively (without gas transfer). The first venting action to raise altitude does not expend power, allowing a different altitude to be interrogated for the second circumnavigation. On the third circumnavigation, the onboard helium pump is used to significantly lower the equilibrium altitude on the dayside when solar power is abundant.

Finally, the longitude of the aerobot continuously cycles as it is blown westward around the planet, and small drifts in the aerobot latitude mean that specific ground features are not necessarily revisited. Depending on the Hadley cell meridional winds that vary with altitude (and are function of the global circulation of the season) \cite{horinouchi2020waves}, the aerobot can expect to slowly spiral poleward or equatorward and can control that motion to the extent that altitude control authority and weather models are available. Furthermore, the aerobot trajectory is expected to be invaluable for refining these weather models, as the aerobot drifts as a Lagragian tracer of the meridional and zonal winds.

The FLOATS simulation routine, now partially validated through the Nevada experimental flights, can accordingly act as a mission planning utility for executing a Venus variable-altitude aerobot flight. Pumping and venting choices can be integrated with power availability tables, and balanced against instrument and telecom power needs, to run a large volume of mission scenarios to explore different desired observation outcomes.

\section{Conclusion \label{sec:conclusion}}
In this paper, we have documented two experimental test flights of a prototype Venus aerobot in the Nevada Black Rock desert and a first-principle dynamics model for understanding its behavior. Our specific contributions are:
\begin{enumerate}    
    \item \textbf{Prototype Fabrication:} We describe our aerobot which is fabricated from Venus-relevant materials --- building upon smaller prototypes in prior work \cite{izraelevitz2022subscale, hall2021prototype} and showing an increase in balloon scale and manufacturing fidelity. The aerobot in this flight test includes a metalized bilaminate of Teflon\textsuperscript{\textregistered} and Kapton\textsuperscript{\textregistered} for the outer ZP envelope, and a Vectran\textsuperscript{\textregistered} woven inner SP envelope with urethane bladder. These materials are the flight materials expected for surviving Venus thermal conditions of $60^\circ$C \cite{seiff1985models}, >94\% concentration sulfuric acid aerosols \cite{cutts2022explore}, and the intense solar heating of 2300 W/m\textsuperscript{2} \cite{robinson2018linearized}.
    \item \textbf{Flight Demonstration in a Relevant Environment:} We show the first flight demonstration of a Venus-capable (i.e. environmentally compatible) altitude controlled aerobot, building capability beyond the constant-altitude VeGa technology. These flights occurred over a kilometer-extent atmospheric density range that is identical to the center of the targeted cloud layer altitudes on Venus (54 - 55~km).
    \item \textbf{Buoyancy Modulation through Helium Gas Transfer:} We demonstrate the ability to move helium gas with a pumping/venting system between balloon chambers to execute these altitude control maneuvers. This adapts the balloon-in-balloon architecture described in \citet{voss2009advances} to a Venus design, protecting the high-pressure envelope from the acid aerosols in the external atmosphere. The buoyancy modulation authority is shown to be sufficient to vary altitude in the aerobot free-flight.
    \item \textbf{Flight Instrumentation:} We describe and implement an instrumentation set for capturing the aerobot dynamics and thermodynamics in sufficient detail for deriving a first-principal dynamic model of the aerobot. This includes barometers, temperature sensors, radiometers, cameras, and GPS trackers; as well as piloting infrastructure from the command center. While this instrumentation does not yet utilize spaceflight-grade components, it provides a template for the Venus avionics of future work.
    \item \textbf{Model Description \& Validation:} We describe models of the aerobot shape, node temperature, and altitude dynamics as a function of the solar/infrared radiative environment, atmospheric temperature/pressure, vertical wind, and gas exchange actuation. We validate these models against the data from the test flights and account for discrepancies between them.
    \item \textbf{Model Application:} We apply this model to Venus flight cases with larger aerobot designs under Venus environmental conditions derived from planetary data in the literature. These Venus flight predictions allow for the tweaking of aerobot designs and can output mission planning parameters such as the altitude, ground track, and solar availability for the aerobot in flight.
\end{enumerate}

The flight dynamics at kilometer-scale altitude range have been demonstrated in this paper, yet other technologies for the Venus aerobot still require further development. These include: (1) building larger aerobot prototypes with sufficient lift capacities for the science described in Section \ref{Sec.Introduction}; (2) continued relevant-environment testing of envelope materials against aerosol particles of concentrated sulfuric acid and full-scale tensile loads; and (3) running flight tests over a greater altitude control capability and longer duration under diverse solar influence conditions. 

Additionally, the deployment of an aerobot from a parachute is another key technology for a Venus mission. While it may be possible to inflate a Venus aerobot from the Venus surface using a much higher temperature-capable envelope, the primary advantage of these aerial platforms is to avoid the harsh Venus ground conditions that challenge landed missions. Practically, the aerobot should therefore be deployed mid-air during atmospheric descent. The packaging for cruise, the extraction of the aerobot to a hanging configuration beneath a parachute, and the subsequent inflation from disposable helium tanks are therefore critical deployment processes that are a prerequisite for the flight phase of the aerobot. This process has been demonstrated for the VeGa balloons \cite{sagdeev1986vega}, and larger JPL constant-altitude balloons have demonstrated most of the process (but not yet the transition to free flight) \cite{hall2011technology}. However, no end-to-end deployment demonstration has been attempted for a variable altitude balloon-in-balloon. An inflation timeline and initial indoor experiments are described in \cite{gatto2024inflation}, but work remains to validate the deployment architecture in a free atmosphere under lateral wind disturbances.

The two test flights described in this manuscript are a key step in the technology development path of Venus aerobots. Although there is work to go before all technologies are ready, prototypes are being manufactured, instrumented, flown, recovered, and reflown --- demonstrating a capability that could open the skies of Venus to the next generation of planetary science investigations. Where rovers have been the key to surface mobility on the Moon and Mars, balloons offer an analogous aerial capability for the second planet.

\pagebreak
\section*{Appendix: Sensor Details and Shape Model Supplement}
\label{sec.Appendix}
\subsection{Avionics and Sensor Calibration}
\label{app:sensorcal}
\begin{figure}
    \centering
    \includegraphics[width=\textwidth]{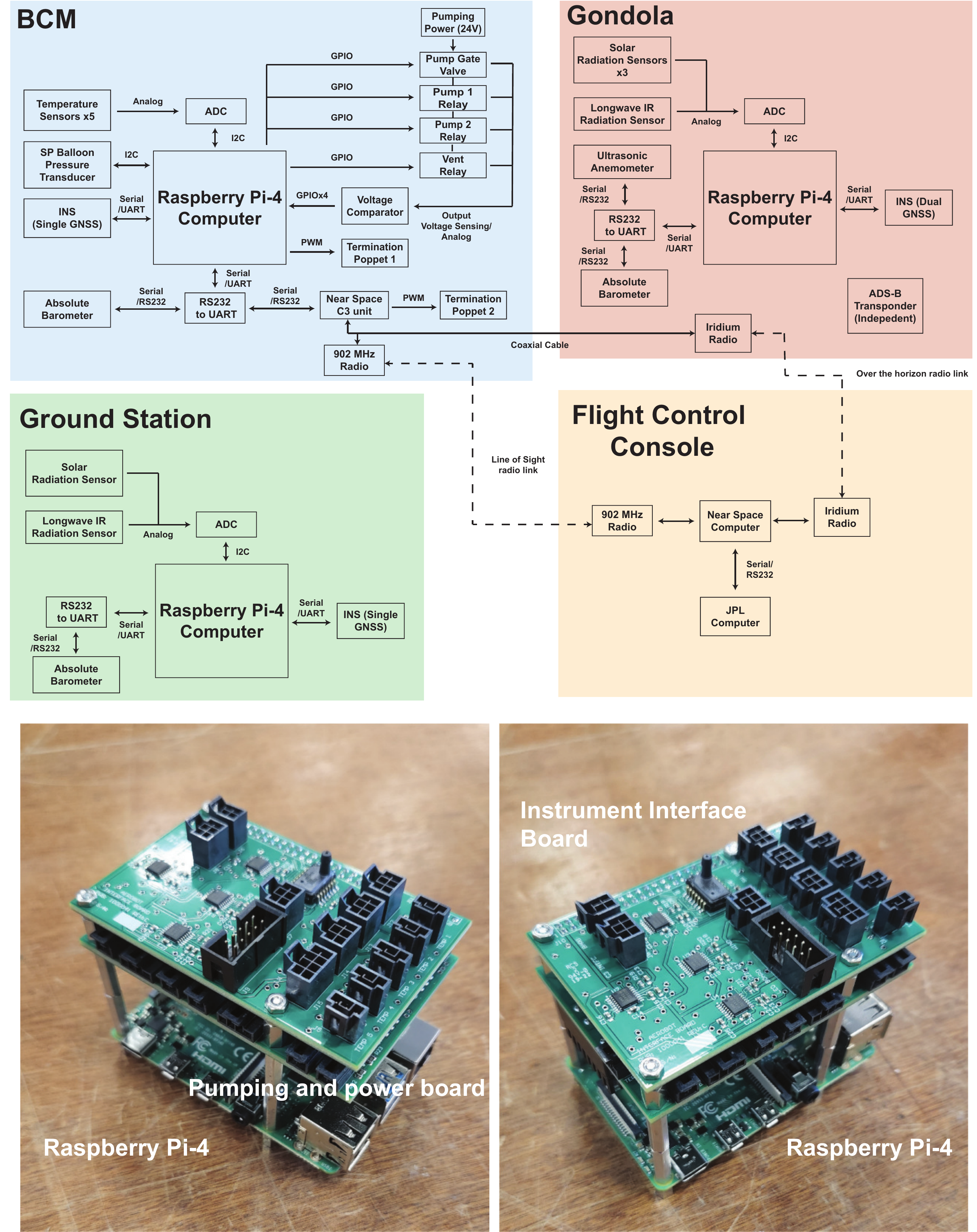}
    \caption{Avionics diagram of the instrumentation and gas transfer system split over four platforms. Images show the common instrumentation stack with the pumping and power board and the instrument interface board.}
    \label{fig:flightstack}
\end{figure}

We developed a custom avionics system to ensure reliable commanding and telemetry from the aerobot. Two stackable custom boards were designed and stacked on a Raspberry Pi computer for the actuation of pumps and vents, and the collection and telemetry of flight data. Identical avionics stacks were used for the BCM, Gondola and Ground platforms to maximize inter-usability and redundancy. Commanding and data transfer between the ground station and the aerobot was achieved through a serial connection with Aerostar's C3 system and redundant 902 MHz and Iridium-based wireless links. Figure \ref{fig:flightstack} shows the avionics diagram the flight commanding and instrumentation system.

\subsection{Instrument Calibration}
\begin{figure} [t]
    \centering
    \includegraphics[width=0.7\textwidth]{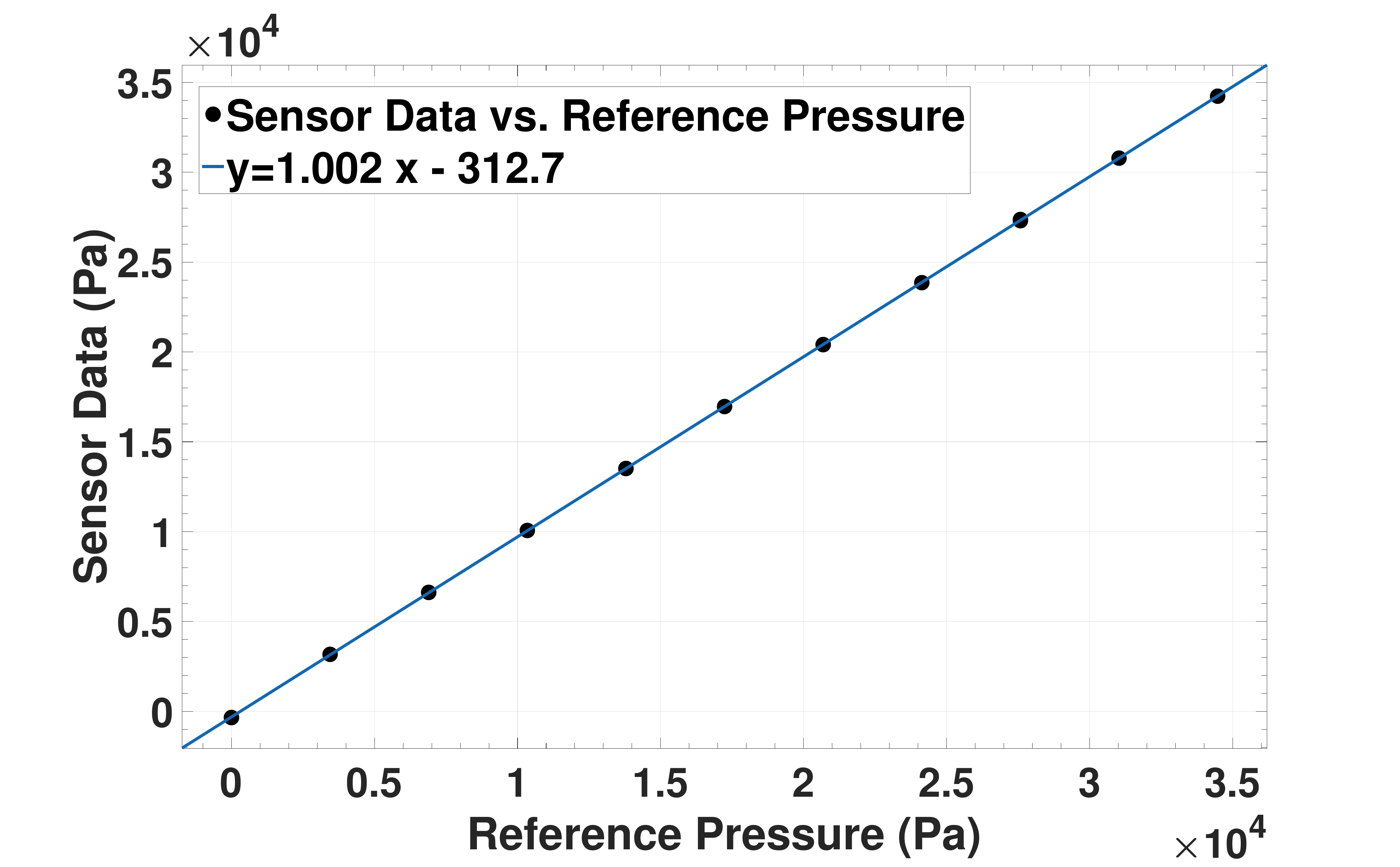}
    \caption{Calibration of the SP transducer. Test pressure was varied in the ascending and descending direction from 0 to 5 psig (approximately 35 kPa). Measurements of pressure by the SP transducer were compared with a NIST-standard barometer. The resulting data points show good agreement between the two barometers.}
    \label{fig:pressureCal}
\end{figure}

\begin{figure}
    \includegraphics[width=\textwidth]{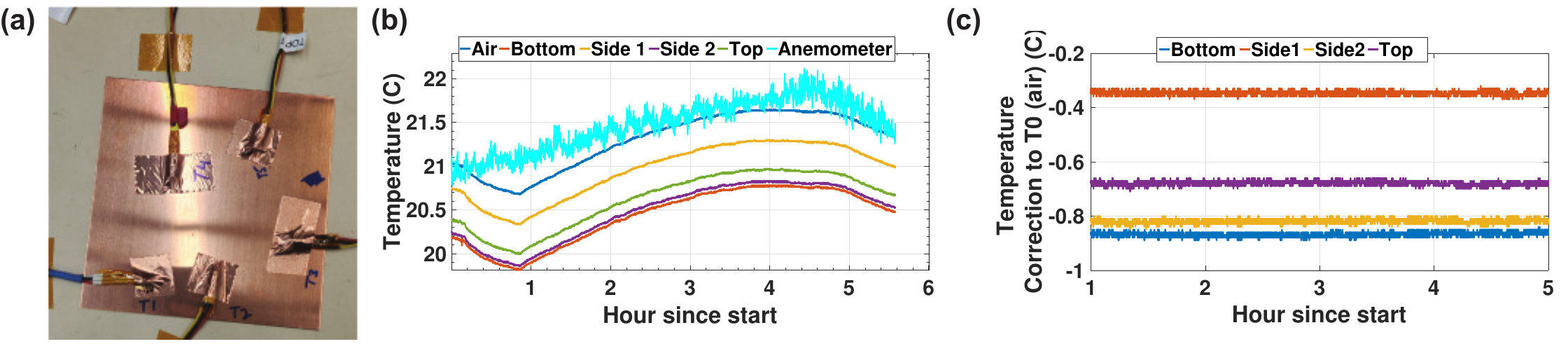}
    \caption{Calibration of the temperature sensors. All sensors were attached to a copper plate using copper tape to ensure good thermal contact. Each sensor had the same length of cable as in the flights. Measurements show relative bias from manufacturing uncertainties and capacitance from the length of cable.}
    \label{fig:thermistorCal}
\end{figure}

\begin{figure}
    \centering
    \includegraphics[width=\textwidth]{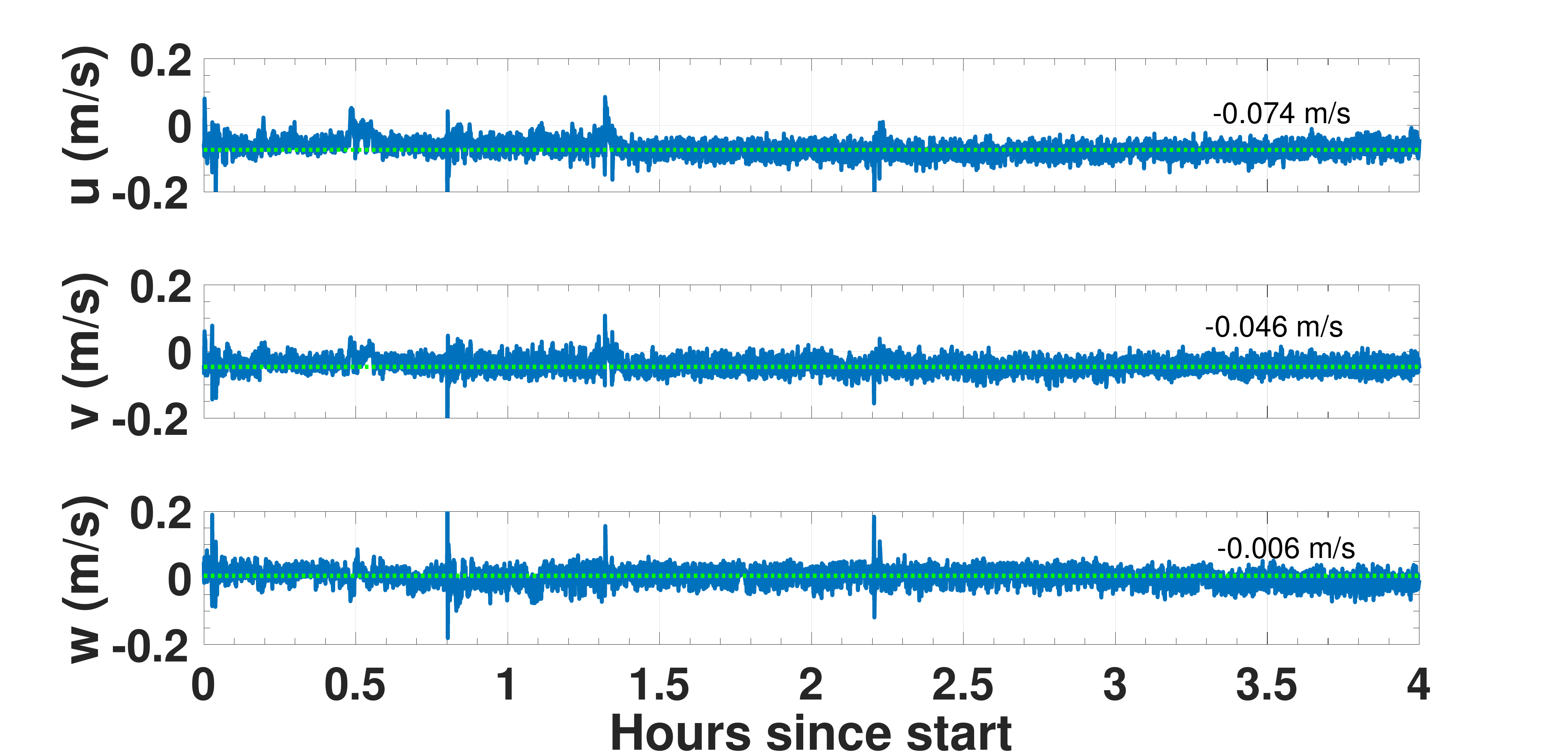}
    \caption{Calibration of the Wind sensor. The ultrasonic anemometer was placed in a windless room and wind speeds were recorded along the three measurement axes of the anemometer. Measurements show very small biases (ignoring brief transients). }
    \label{fig:windCal}
\end{figure}

We performed post-flight calibration of several instruments and incorporated any corrections and biases resulting from the calibration into data recorded during flight. Figures \ref{fig:pressureCal}-\ref{fig:windCal} show results from the calibration of the SP transducer, temperature sensors, and anemometer. The Paroscientific microbarometer and the radiation sensors are supplied pre-calibrated from manufacturers along with a calibration sheet. The SP transducer was calibrated against a NIST-standard barometer by varying the test pressure across the full dynamic range of the SP transducer in the ascending and descending directions. Temperature sensors were thermally coupled with a large copper plate, with the cables in the flight configuration. The temperature of the copper plate was assumed to be uniform and the relative difference between the temperature sensors was noted. Air temperature was also recorded using the ultrasonic anemometer. For anemometer calibrations, the instrument was placed indoors in a windless room. After allowing transients to settle (approximately 90 minutes), the values of wind velocity along the three axes were noted. 


\subsection{Buoyancy force that contributes to initial balloon film stress}\label{app:BuoyancyForce}
\begin{figure}[!ht]
    \centering
    \includegraphics[width=0.5\linewidth]{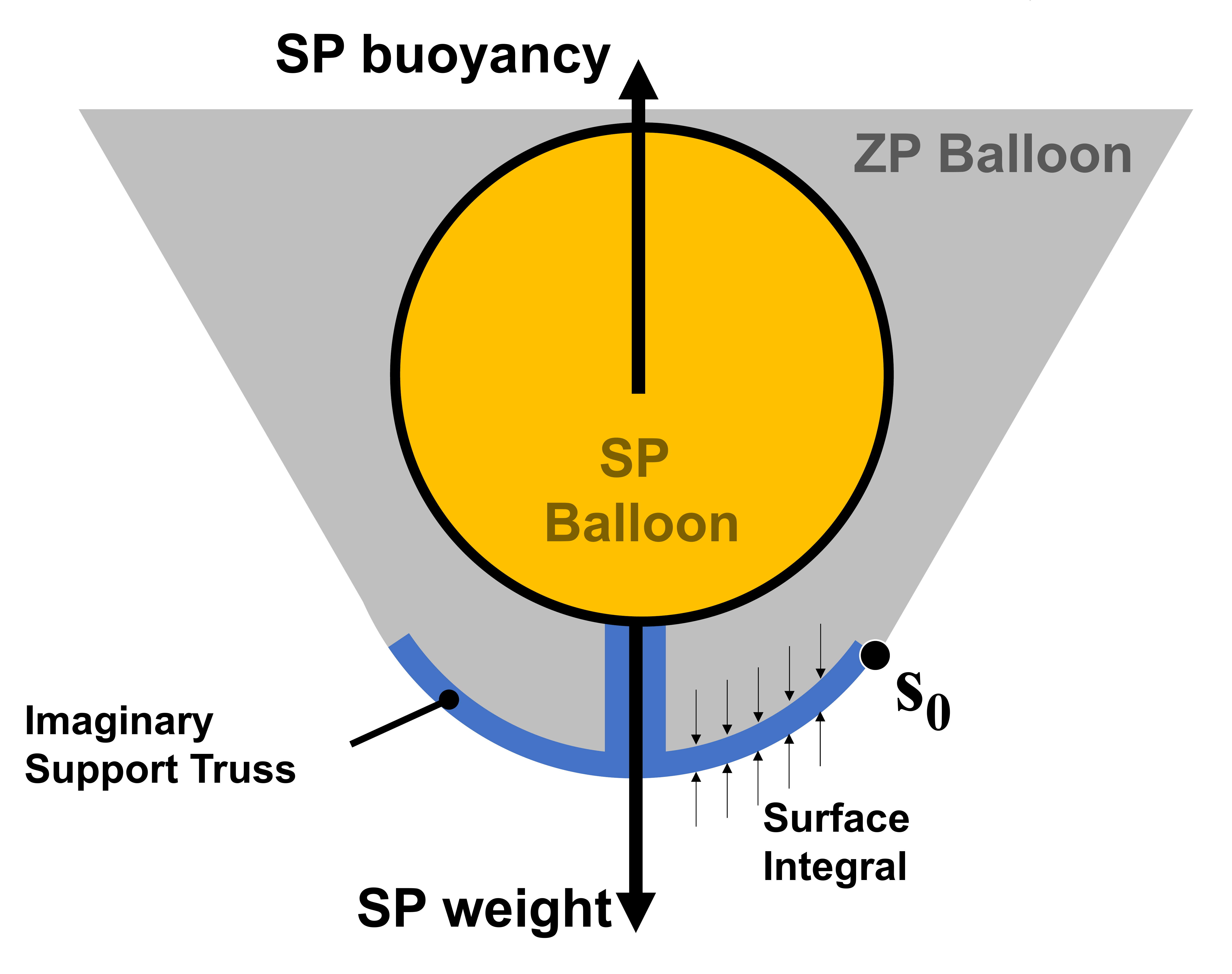}
    \caption{Derivation Diagram for Base Section Buoyancy: The SP weight, minus it's own buoyancy within the ZP balloon, can be imagined to be supported by a support truss structure that creates the bottom spherical shape. The buoyancy of that shape is the given by a surface integral pressure difference.}
    \label{fig:balloonDiagram}
\end{figure}

The purpose of this appendix section is to determine the buoyancy force of the base of the aerobot which is an important step in creating boundary conditions for our balloon shape model (Section \ref{subsubsec:shape}). The initial stress at $s_0$ is hereby derived, staring with the subsystem shown in Figure \ref{fig:balloonDiagram}.

First, we should note that the internal positive superpressure of the SP balloon has no effect, it acts on all radial surfaces equally and should not contribute to the surface pressure distributions (or buoyancy) of the base section. This pressure is instead included in the weight of the SP balloon. However, we can imagine the SP balloon rests upon an unsealed truss structure that allows the ZP helium gas to reach the gap between balloons. The vertical component of the tension on the ZP balloon material at $s_0$ can therefore be given by Eq. \ref{eq:TAF} as the sum of the payload weight, SP weight, and ZP material weight up to $s_0$, minus the SP buoyancy within the ZP balloon and minus the external buoyancy integral up to $s_0$.

To determine the external buoyancy force on the base section of the aerobot, we therefore only need to integrate the difference between the atmospheric pressure and the helium pressure along the contact surface between the balloons. 

Let the small variation of $P_\text{atm}$ with respect to height be given by:
\begin{equation*}
    P_\text{atm} = P_\text{atm,0} - \rho_\text{atm} g z,
\end{equation*}
where for the purposes of this appendix (unlike the rest of the manuscript) $z$ is measured from the center of the SP balloon with positive upward for an easier formulation of the integrals along the SP surface. The small variation in the ZP pressure $P_\text{ZP}$ with height is then given by,
\begin{equation*}
    P_\text{ZP} = P_\text{ZP,0} - \rho_\text{ZP} g z.
\end{equation*}
Then, the pressure difference between the inside and outside of the balloon is given by,
\begin{equation*}
    P_\text{diff} = P_\text{diff,0} - \rho_\text{diff} g z.
\end{equation*}

Let's use spherical coordinates to parameterize the contact area of the ZP and SP balloon envelopes, centered at the SP center,
\begin{align*}
    x &= r \sin(\psi) \cos(\phi)\\
    y &= r \cos(\psi) \sin(\phi)\\
    z &= r \cos(\phi)
\end{align*}
where $\psi \in [0, 2\pi)$ is an angle about the vertical axis and $\phi \in [0,\pi]$ is the angle from the SP apex. In our case, the axisymmetric assumption means that the forces in the $x$ and $y$ directions will be zero, so all we need to worry about are the forces in the $z$ direction. 

The force at any given location created by the pressure difference in the z-direction is,
\begin{equation*}
    F_\text{diff,z} = -\Big(P_\text{diff,0} - \rho_\text{diff} g R_\text{SP} \cos(\phi)\Big)\cos(\phi).
\end{equation*}
Integrating this over the correct portion of the balloon yields,
\begin{align*}
    F_{\text{buoy}}(s_0) &= \int_0^{2\pi} \int_{\pi-\beta}^{\pi} -\Bigg(\Big(P_\text{diff,0} - \rho_\text{diff} g R_\text{SP} \cos(\phi)\Big)\cos(\phi)\Bigg)\Big(R_\text{SP}^2 \sin(\phi)\Big) \dd\phi \dd\psi\\
    F_{\text{buoy}}(s_0) &= -\int_0^{2\pi} \int_{\pi-\beta}^{\pi} \Big(P_\text{diff,0} \cos(\phi)\Big)\Big(R_\text{SP}^2 \sin(\phi)\Big) \dd\phi \dd\psi + \\&\quad \int_0^{2\pi} \int_{\pi-\beta}^{\pi} \Big(\rho_\text{diff} g R_\text{SP} \cos^2(\phi)\Big)\Big(R_\text{SP}^2 \sin(\phi)\Big) \dd\phi \dd\psi\\
    F_{\text{buoy}}(s_0) &= -P_\text{diff,0}R_\text{SP}^2 \int_0^{2\pi} \int_{\pi-\beta}^{\pi} \cos(\phi)\sin(\phi) \dd\phi \dd\psi + \\&\quad +\rho_\text{diff} g R_\text{SP}^3 \int_0^{2\pi} \int_{\pi-\beta}^{\pi} \cos^2(\phi)\sin(\phi) \dd\phi \dd\psi\\
    F_{\text{buoy}}(s_0) &= P_\text{diff,0}R_\text{SP}^2 (2\pi)\frac{\sin^2(\beta)}{2}+\rho_\text{diff} g R_\text{SP}^3 (2\pi)\frac{1-\cos^3(\beta)}{3}\\
    F_{\text{buoy}}(s_0) &= \pi \sin^2(\beta) P_\text{diff,0}R_\text{SP}^2 +\frac{2\pi}{3}\Big(1-\cos^3(\beta)\Big)\rho_\text{diff} g R_\text{SP}^3
\end{align*}

The value of $P_\text{diff,0}$ can be calculated from the point where internal and external pressures are equal, $z_{p0}$, that we solve for in the BVP. These two are related via:
\begin{equation*}
    P_\text{diff,0} = g\rho_\text{diff}\Big(z_{p0}-R_\text{SP}\Big).
\end{equation*}
Substituting this into the solution for $F_{\text{buoy}}(s_0)$ yields,
\begin{equation*}
    \boxed{F_{\text{buoy}}(s_0) = \pi \sin^2(\beta) g\rho_\text{diff}\Big(z_{p0}-R_\text{SP}\Big)R_\text{SP}^2 +\frac{2\pi}{3}\Big(1-\cos^3(\beta)\Big)\rho_\text{diff} g R_\text{SP}^3}
\end{equation*}

As a validation check, we can evaluate $F_{\text{buoy}}(s_0)$ for $\beta = \pi$ and we should get the volume of a sphere multiplied by the density difference and acceleration due to gravity, i.e., $ \frac{4}{3}\pi R_\text{SP}^3 \rho_\text{diff} g$.
\begin{align*}
    F_{\text{buoy}}(s_0)\Big|_{\beta = \pi} &= 0 + \frac{2\pi}{3}\Big(1-(-1)^3\Big)\rho_\text{diff} g R_\text{SP}^3\\
    F_{\text{buoy}}(s_0)\Big|_{\beta = \pi} &= \frac{4}{3}\pi\rho_\text{diff} g R_\text{SP}^3
\end{align*}

\section*{Acknowledgments}
The research described in this paper was funded by the Jet Propulsion Laboratory, California Institute of Technology, under contract 80NM0018D0004  with the National Aeronautics and Space Administration. The authors would like to thank: Kirk Barrow, Charlene Paloma, Tanya Samra, Faustino Chirino, and Lance Christensen for making our field tests a reality; Ahyde Lara, Richard Bauer, and Allen Dial for supporting the aerobot fabrication and flight; the Black Rock Nevada BLM Office and the Tillamook Air Museum for the use of their property; Satish Khanna, Chris Yahnker, Pat Beauchamp, Kim Aaron, Peter Dillon, Dara Sabahi, Michael Meacham, John Gallon, Katie Siegel, Michael Schein, and Jim Denman for their technical guidance of the program; Bill Warner, Dominic Aldi, Jared Leblanc, and Emma Bradford for the material testing of balloon envelopes; Spencer Diehl for reprocessing FLOATS simulations after first review, and Blair Emanuel and Tonya Beatty for the post-flight evaluation of the aerobot prototype after the Nevada field campaign.

Reference herein to any specific commercial product, process, or service by trade name, trademark, manufacturer, or otherwise, does not constitute or imply its endorsement by the United States Government or the Jet Propulsion Laboratory, California Institute of Technology
\bibliography{venusAeroRefs}

\begin{thebibliography}{37}
\newcommand{\enquote}[1]{``#1''}
\providecommand{\natexlab}[1]{#1}
\providecommand{\url}[1]{\texttt{#1}}
\providecommand{\urlprefix}{URL }
\expandafter\ifx\csname urlstyle\endcsname\relax
  \providecommand{\doi}[1]{\discretionary{}{}{}https://doi.org/#1}\else
  \providecommand{\doi}[1]{\discretionary{}{}{}\urlstyle{rm}\url{https://doi.org/#1}}\fi

\bibitem[{Marov and Huntress(2011)}]{marov2011soviet}
Marov, M.~Y., and Huntress, W., \emph{Soviet Robots in the Solar System.
  Mission technologies and discoveries}, Springer, 2011.
\newblock \urlprefix\url{https:doi.org/10.1007/978-1-4419-7898-1_3}.

\bibitem[{Balint and Baines(2008)}]{balint2008nuclear}
Balint, T., and Baines, K., \enquote{Nuclear polar VALOR: an ASRG-enabled Venus
  balloon mission concept,} \emph{AGU Fall Meeting Abstracts}, Vol. 2008, 2008,
  pp. P33A--1439.

\bibitem[{Cutts et~al.(2018)Cutts, Matthies, and Thompson}]{cutts2018venus}
Cutts, J.~A., Matthies, L., and Thompson, T.~W., \enquote{Venus Aerial
  Platforms Study,} Tech. Rep. JPL D-102569, October 2018.
\newblock
  \urlprefix\url{https://science.nasa.gov/resource/aerial-platforms-for-the-scientific-exploration-of-venus/}.

\bibitem[{LLC(2021)}]{loon2021}
LLC, L., \emph{Loon Library: Lessons from Building Loon’s Stratospheric
  Communications Service}, 2021.
\newblock \urlprefix\url{https://x.company/projects/loon/the-loon-collection/}.

\bibitem[{Hall et~al.(2021)Hall, Pauken, Schutte, Krishnamoorthy, Aiazzi,
  Izraelevitz, Lachenmeier, and Turner}]{hall2021prototype}
Hall, J.~L., Pauken, M., Schutte, A., Krishnamoorthy, S., Aiazzi, C.,
  Izraelevitz, J., Lachenmeier, T., and Turner, C., \enquote{Prototype
  Development of a Variable Altitude Venus Aerobot,} \emph{AIAA Paper
  2021-2696, presented at the AIAA Aviation Forum}, 2021.
\newblock \urlprefix\url{https://doi.org/10.2514/6.2021-2696}.

\bibitem[{{National Oceanic and Atmospheric
  Administration}(1976)}]{national1976us}
{National Oceanic and Atmospheric Administration}, \enquote{US Standard
  Atmosphere, 1976,} \emph{Technical Report}, , No. S/T 76-1562, 1976.
\newblock
  \urlprefix\url{https://www.ngdc.noaa.gov/stp/space-weather/online-publications/miscellaneous/us-standard-atmosphere-1976/us-standard-atmosphere_st76-1562_noaa.pdf}.

\bibitem[{Seiff et~al.(1985)Seiff, Schofield, Kliore, Taylor, Limaye,
  Revercomb, Sromovsky, Kerzhanovich, Moroz, and Marov}]{seiff1985models}
Seiff, A., Schofield, J., Kliore, A., Taylor, F., Limaye, S., Revercomb, H.,
  Sromovsky, L., Kerzhanovich, V., Moroz, V., and Marov, M.~Y., \enquote{Models
  of the structure of the atmosphere of Venus from the surface to 100
  kilometers altitude,} \emph{Advances in Space Research}, Vol.~5, No.~11,
  1985, pp. 3--58.
\newblock \urlprefix\url{https://doi.org/10.1016/0273-1177(85)90197-8}.

\bibitem[{{Venus Exploration Analysis Group (VEXAG)}(2019)}]{vexag2019}
{Venus Exploration Analysis Group (VEXAG)}, \enquote{2019 Goals, Objectives,
  and Investigations,} , 2019.
\newblock
  \urlprefix\url{https://www.lpi.usra.edu/vexag/documents/reports/VEXAG_Venus_GOI_2019.pdf}.

\bibitem[{{National Academies of Sciences, Engineering, and
  Medicine}(2022)}]{national2022origins}
{National Academies of Sciences, Engineering, and Medicine}, \enquote{Origins,
  Worlds, and Life: A Decadal Strategy for Planetary Science and Astrobiology
  2023-2032,} 2022.
\newblock
  \urlprefix\url{https://nap.nationalacademies.org/catalog/26522/origins-worlds-and-life-a-decadal-strategy-for-planetary-science}.

\bibitem[{Gilmore et~al.(2020)Gilmore, Beauchamp, Lynch, and
  Amato}]{gilmore2020venus}
Gilmore, M., Beauchamp, P., Lynch, R., and Amato, M., \enquote{Venus Flagship
  Mission Decadal Study Final Report,} \emph{A Planetary Mission Concept Study
  Report Presented to the Planetary and Astrobiology Decadal Survey}, 2020.
\newblock
  \urlprefix\url{https://science.nasa.gov/wp-content/uploads/2023/11/venus-flagship-mission.pdf}.

\bibitem[{O’Rourke et~al.(2021)O’Rourke, Jessup, Lynch, Kasting, Bernard,
  Navarro, and Amato}]{orourke2021advents}
O’Rourke, J., Jessup, K., Lynch, R., Kasting, J., Bernard, M., Navarro, T.,
  and Amato, M., \enquote{ADVENTS: Assessment and Discovery of Venus’ past
  Evolution and Near-Term climatic and geophysical State,} \emph{Mission
  Concept Study Report to the NRC Planetary Science and Astrobiology Decadal
  Survey 2023–2032}, NASA Goddard Space Flight Center, Green Bank, Maryland,
  2021.
\newblock
  \urlprefix\url{https://science.nasa.gov/wp-content/uploads/2023/10/advents-venus-orbiter-aerobot-and-dropsonde.pdf}.

\bibitem[{Cutts et~al.(Submitted 2022)Cutts, Baines, Byre, Dorsky, Frazier,
  Izraelevitz, O’Rourke, Pauken, Seager, and Wilson}]{cutts2022explore}
Cutts, J., Baines, K., Byre, P., Dorsky, L., Frazier, W., Izraelevitz, J.,
  O’Rourke, J., Pauken, M., Seager, S., and Wilson, C., \enquote{Exploring
  the Clouds of Venus: Science Driven Aerobot Missions to our Sister Planet,}
  \emph{2022 IEEE Aerospace Conference}, IEEE, Submitted 2022.
\newblock \urlprefix\url{https://doi.org/10.1109/AERO53065.2022.9843740}.

\bibitem[{Izraelevitz et~al.(2022)Izraelevitz, Pauken, Krishnamoorthy, Goel,
  Aiazzi, Dorsky, Cutts, Hall, Turner, and
  Lachenmeier}]{izraelevitz2022subscale}
Izraelevitz, J., Pauken, M., Krishnamoorthy, S., Goel, A., Aiazzi, C., Dorsky,
  L., Cutts, J., Hall, J.~L., Turner, C., and Lachenmeier, T.,
  \enquote{Subscale Prototype and Hangar Test Flight of a Venus
  Variable-Altitude Aerobot,} \emph{2022 IEEE Aerospace Conference (AERO)},
  IEEE, 2022, pp. 1--11.
\newblock \urlprefix\url{https://doi.org/10.1109/AERO53065.2022.9843453}.

\bibitem[{Voss(2009)}]{voss2009advances}
Voss, P., \enquote{Advances in Controlled Meteorological (CMET) balloon
  systems,} \emph{AIAA Balloon Systems Conference}, 2009, p. 2810.
\newblock \urlprefix\url{https://doi.org/10.2514/6.2009-2810}.

\bibitem[{Hall et~al.(2019)Hall, Cameron, Pauken, Izraelevitz, Dominguez, and
  Wehage}]{hall2019altitude}
Hall, J.~L., Cameron, J., Pauken, M., Izraelevitz, J., Dominguez, M.~W., and
  Wehage, K.~T., \enquote{Altitude-controlled light gas balloons for Venus and
  Titan exploration,} \emph{AIAA Paper 2019-3194, presented at the AIAA
  Aviation Forum}, 2019.
\newblock \urlprefix\url{https://doi.org/10.2514/6.2019-3194}.

\bibitem[{de~Jong(2015)}]{de2015venus}
de~Jong, M., \enquote{Venus altitude cycling balloon,} \emph{Venus Science
  Priorities for Laboratory Measurements}, Vol. 1838, 2015, p. 4030.

\bibitem[{Nock et~al.(1995)Nock, Aaron, Jones, McGee, Powell, Yavrouian, and
  Wu}]{nock1995balloon}
Nock, K., Aaron, K., Jones, J., McGee, D., Powell, G., Yavrouian, A., and Wu,
  J., \enquote{Balloon altitude control experiment (ALICE) project,} \emph{11th
  Lighter-than-Air Systems Technology Conference}, 1995, p. 1632.
\newblock \urlprefix\url{https://doi.org/10.2514/6.1995-1632}.

\bibitem[{Hall et~al.(2008)Hall, Fairbrother, Frederickson, Kerzhanovich, Said,
  Sandy, Ware, Willey, and Yavrouian}]{hall2008prototype}
Hall, J., Fairbrother, D., Frederickson, T., Kerzhanovich, V., Said, M., Sandy,
  C., Ware, J., Willey, C., and Yavrouian, A., \enquote{Prototype design and
  testing of a Venus long duration, high altitude balloon,} \emph{Advances in
  space research}, Vol.~42, No.~10, 2008, pp. 1648--1655.
\newblock \urlprefix\url{https://doi.org/10.1016/j.asr.2007.03.017}.

\bibitem[{Hall et~al.(2011)Hall, Yavrouian, Kerzhanovich, Fredrickson, Sandy,
  Pauken, Kulczycki, Walsh, Said, and Day}]{hall2011technology}
Hall, J., Yavrouian, A., Kerzhanovich, V., Fredrickson, T., Sandy, C., Pauken,
  M.~T., Kulczycki, E.~A., Walsh, G.~J., Said, M., and Day, S.,
  \enquote{Technology development for a long duration, mid-cloud level Venus
  balloon,} \emph{Advances in Space Research}, Vol.~48, No.~7, 2011, pp.
  1238--1247.
\newblock \urlprefix\url{https://doi.org/10.1016/j.asr.2011.05.034}.

\bibitem[{Gatto et~al.(2024)Gatto, Izraelevitz, Pauken, Goel, Lam, and
  Hall}]{gatto2024inflation}
Gatto, V.~L., Izraelevitz, J.~S., Pauken, M.~T., Goel, A., Lam, R., and Hall,
  J.~L., \enquote{Inflation sequence tradeoffs and laboratory demonstration of
  a prototype variable-altitude venus aerobot,} \emph{Acta Astronautica}, Vol.
  214, 2024, pp. 757--773.
\newblock \urlprefix\url{https://doi.org/10.1016/j.actaastro.2023.11.035}.

\bibitem[{{National Centers for Environmental Prediction}(2024)}]{gfs}
{National Centers for Environmental Prediction}, \enquote{Global Forecast
  System (GFS),} , 2024.
\newblock
  \urlprefix\url{https://www.ncei.noaa.gov/products/weather-climate-models/global-forecast}.

\bibitem[{win(2025)}]{windy2025}
\enquote{Windy.com,} , 2025.
\newblock \urlprefix\url{Windy.com}.

\bibitem[{{JPL Robotics Section}(2024)}]{darts}
{JPL Robotics Section}, \enquote{DARTS Homepage,} , 2024.
\newblock
  \urlprefix\url{https://www-robotics.jpl.nasa.gov/how-we-do-it/facilities/the-darts-simulation-laboratory/}.

\bibitem[{Leake(2021)}]{LeakeDissertation}
Leake, C., \enquote{{The Multivariate Theory of Functional Connections: An
  $n$-Dimensional Constraint Embedding Technique Applied to Partial
  Differential Equations},} , 2021.
\newblock \urlprefix\url{https://doi.org/10.48550/arXiv.2105.07070}.

\bibitem[{Leake and Johnston(2021)}]{TfcGithub}
Leake, C., and Johnston, H., \enquote{{TFC: A Functional Interpolation
  Framework},} , 2021.
\newblock \urlprefix\url{https://github.com/leakec/tfc}.

\bibitem[{Leake et~al.(2022)Leake, Johnson, and Mortari}]{leake2022theory}
Leake, C., Johnson, H., and Mortari, D., \emph{The Theory of Functional
  Connections: A Functional Interpolation Framework with Applications}, Lulu.
  com, 2022.

\bibitem[{Carlson and Horn(1983)}]{horne}
Carlson, L.~A., and Horn, W.~J., \enquote{New thermal and trajectory model for
  high-altitude balloons,} \emph{Journal of Aircraft}, Vol.~20, No.~6, 1983,
  pp. 500--507.
\newblock \urlprefix\url{https://doi.org/10.2514/3.44900}.

\bibitem[{Newman(2018)}]{newman2018marine}
Newman, J.~N., \emph{Marine hydrodynamics}, The MIT press, 2018.

\bibitem[{Sarpkaya(2010)}]{sarpkaya2010wave}
Sarpkaya, T., \emph{Wave forces on offshore structures}, Cambridge University
  Press, 2010.

\bibitem[{Hoerner(1958)}]{hoerner1958fluid}
Hoerner, S.~F., \enquote{Fluid-dynamic drag.} \emph{CHAPTER V-Drag of Surface
  Imperfections}, 1958.

\bibitem[{Kreith and Kreider(1974)}]{kreith1974numerical}
Kreith, F., and Kreider, J.~F., \enquote{Numerical prediction of the
  performance of high altitude balloons,} Tech. Rep. NCAR-TN/STR-6, 1974.

\bibitem[{Lienhard and Lienhard(2024)}]{lienhard2024heat}
Lienhard, J.~H., IV, and Lienhard, J.~H., V, \emph{A Heat Transfer Textbook},
  6\textsuperscript{th} ed., Phlogiston Press, Cambridge, MA, 2024.
\newblock \urlprefix\url{https://ahtt.mit.edu}, version 6.00.

\bibitem[{White and Hoffman(2023)}]{white2023earth}
White, P., and Hoffman, J., \enquote{Earth Global Reference Atmospheric Model
  (Earth-GRAM): User Guide,} Tech. Rep. TM-20230014404, 2023.

\bibitem[{Lebonnois et~al.(2016)Lebonnois, Sugimoto, and
  Gilli}]{lebonnois2016wave}
Lebonnois, S., Sugimoto, N., and Gilli, G., \enquote{Wave analysis in the
  atmosphere of Venus below 100-km altitude, simulated by the LMD Venus GCM,}
  \emph{Icarus}, Vol. 278, 2016, pp. 38--51.
\newblock \urlprefix\url{https://doi.org/10.1016/j.icarus.2016.06.004}.

\bibitem[{Robinson and Crisp(2018)}]{robinson2018linearized}
Robinson, T.~D., and Crisp, D., \enquote{Linearized Flux Evolution (LiFE): A
  technique for rapidly adapting fluxes from full-physics radiative transfer
  models,} \emph{Journal of Quantitative Spectroscopy and Radiative Transfer},
  Vol. 211, 2018, pp. 78--95.
\newblock \urlprefix\url{https://doi.org/10.1016/j.jqsrt.2018.03.002}.

\bibitem[{Horinouchi et~al.(2020)Horinouchi, Hayashi, Watanabe, Yamada,
  Yamazaki, Kouyama, Taguchi, Fukuhara, Takagi, Ogohara, Murakami, Peralta,
  Limaye, Imamura, Nakamura, Sato, and Satoh}]{horinouchi2020waves}
Horinouchi, T., Hayashi, Y.-Y., Watanabe, S., Yamada, M., Yamazaki, A.,
  Kouyama, T., Taguchi, M., Fukuhara, T., Takagi, M., Ogohara, K., Murakami,
  S.-Y., Peralta, J., Limaye, S., Imamura, T., Nakamura, M., Sato, T.~M., and
  Satoh, T., \enquote{How waves and turbulence maintain the super-rotation of
  Venus’ atmosphere,} \emph{Science}, Vol. 368, No. 6489, 2020, pp. 405--409.
\newblock \urlprefix\url{https://doi.org/10.1126/science.aaz4439}.

\bibitem[{Sagdeev et~al.(1986)Sagdeev, Linkin, Blamont, and
  Preston}]{sagdeev1986vega}
Sagdeev, R., Linkin, V., Blamont, J., and Preston, R., \enquote{The VEGA Venus
  balloon experiment,} \emph{Science}, Vol. 231, No. 4744, 1986, pp.
  1407--1408.
\newblock \urlprefix\url{https://doi.org/10.1126/science.231.4744.1407}.

\end{thebibliography}

\end{document}